\documentclass[10pt,twocolumn,letterpaper]{article}

\usepackage[pagenumbers]{cvpr} 

\usepackage[dvipsnames]{xcolor}
\usepackage{graphicx}
\usepackage{booktabs}
\usepackage{multirow}
\usepackage{enumitem}
\usepackage{pifont}
\usepackage{wrapfig}
\usepackage{caption}

\definecolor{cvprblue}{rgb}{0.21,0.49,0.74}
\usepackage[pagebackref,breaklinks,colorlinks,citecolor=cvprblue]{hyperref}

\def\paperID{***} 
\def\confName{3DV\xspace}
\def\confYear{2025\xspace}

\title{
Online 3D Scene Reconstruction Using Neural Object Priors
}

\author{
Thomas Chabal${}^{\dagger\ddagger}$
\and
Shizhe Chen${}^{\dagger\ddagger}$
\and
Jean Ponce${}^{\ddagger*}$
\and
Cordelia Schmid${}^{\dagger\ddagger}$
\thanks{
${}^{\dagger}$ Inria.
${}^{\ddagger}$ Department of Computer Science, \'Ecole normale supérieure (ENS-PSL, CNRS, Inria).
${}^{*}$ Courant Institute of Mathematical Sciences and Center for Data Science, New York University.
{\tt\small \{firstname.lastname\}@inria.fr}
}
}

\makeatletter
\def\thanks#1{\protected@xdef\@thanks{\@thanks
        \protect\footnotetext{#1}}}
\makeatother

\newcommand{\cmark}{\ding{51}}%
\newcommand{\xmark}{\ding{55}}%

\begin{document}
\maketitle

\begin{abstract}
This paper addresses the problem of reconstructing a scene online at the level of objects given an RGB-D video sequence. 
While current object-aware neural implicit representations hold promise, they are limited in online reconstruction efficiency and shape completion.
Our main contributions to alleviate the above limitations are twofold.
First, we propose a feature grid interpolation mechanism to continuously update grid-based object-centric neural implicit representations as new object parts are revealed.
Second, we construct an object library with previously mapped objects in advance and leverage the corresponding shape priors to initialize geometric object models in new videos, subsequently completing them with novel views as well as synthesized past views to avoid losing original object details.
Extensive experiments on synthetic environments from the Replica dataset, real-world ScanNet sequences and videos captured in our laboratory demonstrate that our approach outperforms state-of-the-art neural implicit models for this task in terms of reconstruction accuracy and completeness.

\end{abstract}

\section{Introduction}
\label{sec:intro}
Reconstructing a scene from a moving camera is a fundamental task in computer vision, with a wide range of applications in augmented reality~\cite{newcombe2011dtam}, robotics~\cite{adamkiewicz2022navigation,liu2024okrobot}, autonomous driving~\cite{huang2023tpvformer,yang2023unisim}, entertainment or cultural heritage preservation. 
Despite the availability of numerous approaches~\cite{placed2023survey} to tackle the problem, it is still challenging to construct on the fly detailed scene representations made of separate object models  with manageable storage and computation resources.

The prevailing methods in scene reconstruction typically treat the entire scene as a single entity~\cite{curless1996tsdf,newcombe2011dtam,newcombe2011kinectfusion,dai2017bundlefusion,sucar2021imap,zhu2022niceslam,rosinol2022nerfslam,muller2022ingp,chen2022tensorf,wang2023coslam,johari2023eslam}, using for example voxel grids~\cite{curless1996tsdf,newcombe2011dtam,newcombe2011kinectfusion,dai2017bundlefusion}, point clouds~\cite{engel2014lsdslam,engel2016dso}, surfel maps~\cite{whelan2015elasticfusion,runz2017cofusion} or meshes~\cite{rosinol2021kimera} to represent the scene.
Recently, neural implicit representations, \eg, NeRFs~\cite{mildenhall2020nerf}, and Gaussian splatting~\cite{Drett23} have given promising results in scene reconstruction~\cite{sucar2021imap,zhu2022niceslam} thanks to their reduced memory footprint and enhanced accuracy at fine-grained resolutions.
However, extracting individual objects from scene-level implicit representations proves to be time and computation intensive~\cite{blomqvist2023implicitvisionlanguagefeaturefields,kerr2023lerf}, resulting in inflexibility to independently manipulate, modify or replace objects in the scene. 

\begin{figure}[t]
    \centering
    \includegraphics[width=\linewidth]{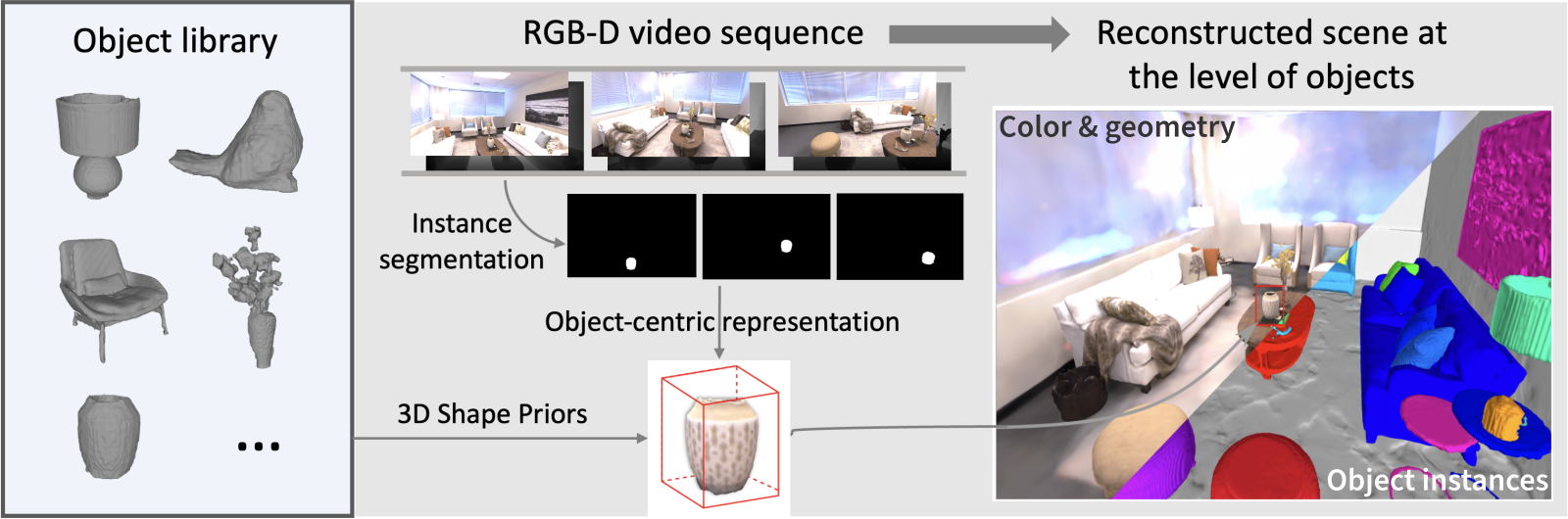}
    \vspace{-1.5em}
    \caption{\small{
    Our method reconstructs scenes at the level of objects from RGB-D videos on the fly. We  leverage 3D shape priors from a pre-computed object library to enhance accuracy and completeness of geometry reconstruction for individual objects.
    }}
    \label{fig:teaser}
    \vspace{-0.5em}
\end{figure}

To address these limitations, there is a growing focus on object-centric neural implicit representations~\cite{kong23vmap,han2023romap} for scene reconstruction.
vMAP~\cite{kong23vmap} is a state-of-the-art approach in this direction that optimizes a separate multi-layer perceptron (MLP) for each object. 
While vMAP is lightweight and operates in an online setting, its ability to reconstruct intricate object details is constrained by its simple MLP model.
RO-MAP~\cite{han2023romap} enhances MLPs with multi-resolution hash coding~\cite{muller2022ingp}.
However, it requires to accumulate all recorded views of an object to estimate the object bounding box for optimization and, thus, sacrifices online efficiency for accuracy.
Furthermore, the above approaches optimize models from scratch without leveraging 3D priors, thus focusing only on object parts visible in the current video and leading to slower and incomplete reconstructions.
Yet, there is a clear benefit to having access to prior object shape knowledge, \ie, we can better reconstruct an object if we have seen similar objects or the same object from different angles before.
Most prior endeavours incorporating object priors either only estimate object poses and cannot reconstruct novel objects~\cite{xiang2018posecnn,li2018deepim,labbe2020cosypose,labbe2022megapose} (see, however~\cite{rothganger20033dobjmodel,Roth07}) or rely on a single latent vector in initialization~\cite{sucar2020nodeslam,zou2022objectfusion} which hardly captures fine-grained object shape and may lose details from past views.

In this work, we overcome the aforementioned limitations with object-centric neural implicit representations for online scene reconstruction.
Given an RGB-D video, we obtain object masks and camera poses at each frame following prior work~\cite{kong23vmap}, then reconstruct all objects in the scene on the fly by optimizing the implicit object representation with differentiable volume rendering.
Each object representation consists of a feature grid tightly encompassing the object of interest and a small MLP to predict occupancy and RGB colors.
Geometry and appearance are disentangled to allow the optimization of one without deteriorating the other.
To allow full online optimization as opposed to the partially online approach in~\cite{han2023romap}, we continuously expand the feature grid and adapt our representation with feature interpolation as unseen parts of objects become visible. 
This enables us to start reconstructing objects as soon as they enter the camera's field of view.
Moreover, we introduce a novel approach to leverage previously observed objects and enhance online optimization of object shapes. 
Specifically, we first construct an object library containing fitted object models from either full 3D meshes or previous mapping videos.
If and when an object is identified in the video, we retrieve the most similar object from the library and align it to the current scene. 
If correctly matched, the shape representation of the retrieved object serves as initialization for the new one.
To ensure keeping the geometric information in the prior object model, we propose to render images from the prior model as a surrogate for past observations and use them as additional signal when optimizing the current model.
We conduct extensive experiments on synthetic environments from the Replica dataset, real-world ScanNet sequences and our own videos. 
Experimental results demonstrate the effectiveness of the proposed system in achieving more accurate and more complete object-aware scene reconstruction.

To summarize, our contribution is two-fold: 
\begin{itemize}[itemsep=0.1em,parsep=0.2em,topsep=0em,partopsep=0em]
    \item We adapt an object-centric grid-based neural implicit representation that we equip with a feature grid interpolation scheme to incrementally reconstruct objects as soon as they become visible. 
    We design it for flexible reuse of geometry across videos, high accuracy and efficiency of optimization.
    \item We leverage pre-constructed shape representations from completely or partially observed objects to accelerate shape optimization and enhance completeness. 
    We address the problem of losing shape details from past views by rendering images from previous viewpoints as additional supervision.
\end{itemize}
The proposed system achieves state-of-the-art performance on both the synthetic Replica dataset, real environments in ScanNet and videos recorded in our laboratory. 
Our code and models are publicly released~\footnote{{\url{https://github.com/thomaschabal/online-scene-reconstruction}}}.

\section{Related Work}
\label{sec:related}

\vspace{-0.5em}
\subsection{Online dense mapping}
\label{sec:rw-reconstruction}
\vspace{-0.5em}
Reconstructing a 3D scene from a set of images is a longstanding problem in computer vision~\cite{curless1996tsdf,hartley2004multiviewgeom}. 
A large branch of the community tackled it with approaches running in an offline fashion, with all the images initially given as input~\cite{rothganger20033dobjmodel,hartley2004multiviewgeom,furukawa2007stereopsis,schoenberger2016sfm,azinovic22surface}. 
Working in an online setting, visual Simultaneous Localization And Mapping (vSLAM) estimates camera poses while mapping scenes from a stream of images, the two problems being tightly linked~\cite{thrun2005probabilisticrobotics}.
While the first real-time approaches considered sparse visual anchors~\cite{davison2007monoslam,klein2007ptam}, later works improved them with better suited features~\cite{murartal2015orbslam,forster2017svo} or semi-dense point clouds~\cite{engel2014lsdslam,engel2016dso}, improving performances while typically operating at frame-rate on a CPU. 
More recent works introduced learning-based approaches for monocular systems~\cite{yang2018dvso,czarnowski2020deepfactors,yang2020d3vo,koestler2021tandem,teed2021droidslam}, but these require heavy computation and tend to not transfer well to scenes outside the training distribution.
With the emergence of consumer-grade depth cameras, dense SLAM systems~\cite{newcombe2011dtam,newcombe2011kinectfusion,schops20153dmodelingonthego,dai2017bundlefusion,rosinol2021kimera} have been developed, typically relying on truncated signed distance functions (TSDF)~\cite{curless1996tsdf,niessner2013hash}, a voxel-based representation of the world that allows for fast incremental merge of depth views.
While the memory usage of these grids grows cubically with the spatial resolution, some works~\cite{huang2021difusion} have explored the storage of latent codes in grids, decoded by a small neural network, to improve reconstruction with coarser grids.
Other scene representations include point clouds~\cite{pizzoli2014remode} and surfels~\cite{whelan2015elasticfusion,runz2017cofusion} which are fast to compute but remain discrete, a limitation for a number of downstream applications.
Few works tackle the problem of object-level SLAM~\cite{mccormac2018fusionpp,xu2019midfusion}, possibly in dynamic scenes~\cite{runz2017cofusion,runz2018maskfusion}, representing signed distance functions as either grids~\cite{mccormac2018fusionpp}, octrees~\cite{xu2019midfusion} or surfels~\cite{runz2017cofusion,runz2018maskfusion}, but they do not leverage any prior on the objects they reconstruct.

\vspace{-0.5em}
\subsection{NeRF-based SLAM}
\vspace{-0.5em}
With the advent of differentiable volume rendering~\cite{mildenhall2020nerf}, localization and mapping approaches based on neural implicit models have been proposed~\cite{sucar2021imap,rosinol2022nerfslam,zhu2022niceslam,johari2023eslam,chung2023orbeezslam,sandstrom2023pointslam,kong23vmap,han2023romap,jiang2023h2mapping,wang2023coslam}, benefiting from improvement in novel view synthesis~\cite{chen2022tensorf,fridovichkeil2022plenoxels,muller2022ingp}.
Unlike traditional voxel grids, point clouds and meshes, implicit representations are light, continuous, differentiable and accurate at any resolution.
These works have originally encoded the spatial position of a point with a small neural network~\cite{sucar2021imap,kong23vmap} before storing features in a point cloud~\cite{sandstrom2023pointslam} or on a sparse~\cite{jiang2023h2mapping} or dense~\cite{zhu2022niceslam, johari2023eslam} grid, possibly indexed by a hashing function~\cite{wang2023coslam,rosinol2022nerfslam,chung2023orbeezslam,jiang2023h2mapping,han2023romap}. 
A small MLP is then used as a decoder.
If some systems follow the original NeRFs by modelling the world with density functions, this formulation does not explicitly define surfaces and has thus been replaced by signed distance functions~\cite{wang2021neus,ortiz2022isdf} or, rather equivalently, occupancy values~\cite{oechsle2021unisurf} in later reconstruction approaches.

Prior NeRF-based works mainly consider the world as a single entity and process it with one single model~\cite{sucar2021imap,zhu2022niceslam,johari2023eslam,wang2023coslam,rosinol2022nerfslam,jiang2023h2mapping,chung2023orbeezslam,sandstrom2023pointslam}. 
This makes it difficult to extract objects or reuse objects in different scenes, in particular for representations relying on hash grids or specific positional encodings.
Furthermore, such models uselessly represent empty space in rooms.
Some recent works further exploit instance segmentation masks to isolate and reconstruct objects in real-time~\cite{wen23bundlesdf,kong23vmap,han2023romap}. 
BundleSDF~\cite{wen23bundlesdf} relies on an efficient hash grid representation~\cite{muller2022ingp,niessner2013hash} but is restricted to reconstructing a single object. 
Conversely and closest to our work, vMAP~\cite{kong23vmap} and RO-MAP~\cite{han2023romap} parallelize the optimization of multiple models, one per object. 
However, the former only relies on MLPs, leading to poor accuracy and slow computation time, while the latter exploits a combination of hash grids and MLPs but it requires objects to be fully in view to estimate their geometry.
Other approaches extract accurate object reconstructions by modelling specifically the presence of objects when optimizing a single scene representation~\cite{wu2022objectsdf,wu2023objectsdfpp}, but they run offline.
None of the methods presented here reuses any knowledge from outside the environment they are optimized on.

\vspace{-0.5em}
\subsection{3D reconstruction with object priors}
\vspace{-0.5em}
\label{sec:rw-objects}
Using object priors has been explored extensively in several fields of computer vision.
As an object-level SLAM system, SLAM++~\cite{salasmoreno2013slampp} exploits a database of few objects mapped offline to refine its localization, but does not additionally map novel objects in a scene.
Another set of works reconstruct objects by learning latent codes for shapes or categories with large object databases~\cite{sucar2020nodeslam,runz2020frodo,wang2021dspslam,zou2022objectfusion,shan2021ellipsdf} and optimizing one code per novel object given input views. However, they are unable to map objects of categories unseen at training and often fail to capture shape details visible in input views.
Recent works also leverage diffusion models to generate realistic images for unseen viewpoints of an object~\cite{poole2023dreamfusion,wang2024morpheus,kong2024eschernet}, but these do not represent the actual unseen parts.
In object pose estimation, numerous approaches have been proposed to estimate the 6D pose of an object in one or multiple observations of a scene given a 3D model of that object~\cite{xiang2018posecnn,li2018deepim,labbe2020cosypose,haugaard2022surfemb,labbe2022megapose,ornek2023foundpose,rothganger20033dobjmodel,Roth07}.
Current approaches to that problem typically employ neural networks to estimate a coarse pose~\cite{xiang2018posecnn,labbe2020cosypose,haugaard2022surfemb,ornek2023foundpose} and then apply refinement by rendering the object~\cite{li2018deepim,labbe2022megapose}. 
While these models fit in a reconstruction pipeline, they are usually restricted to estimating poses for the objects seen at training time~\cite{xiang2018posecnn,li2018deepim,labbe2020cosypose,haugaard2022surfemb}, require an accurate 3D model of each object to be available and cannot reconstruct novel objects.

\vspace{-0.9em}
\section{Method}
\label{sec:method}
\vspace{-0.6em}
Given an RGB-D video $V$ of a static scene captured by a moving camera $C$, our goal is to build object-centric representations for the scene that can accurately reconstruct the whole scene at the level of objects.
We assume for that, as in~\cite{han2023romap,kong23vmap}, that we have access at each frame to object masks tracked consistently in the whole video, one mask per object in the scene, and camera poses expressed in a fixed world frame $w$, both provided by external off-the-shelf systems, \eg,~\cite{murartal2015orbslam,murartal2017orbslam2,cheng2022xmem,cheng2023trackinganything}.
In the following, we first describe our object-centric model for online scene reconstruction and its optimization in~\cref{subsec:object-centric}. 
We then introduce in~\Cref{subsec:init-database} the offline construction of an object library and how to leverage it to enhance the online shape optimization of the same objects observed in novel scenes with different viewpoints.

\vspace{-0.5em}
\subsection{Online object-centric scene reconstruction}
\label{subsec:object-centric}
\vspace{-0.25em}

\begin{figure}[t]
    \centering
    \includegraphics[width=0.85\linewidth]{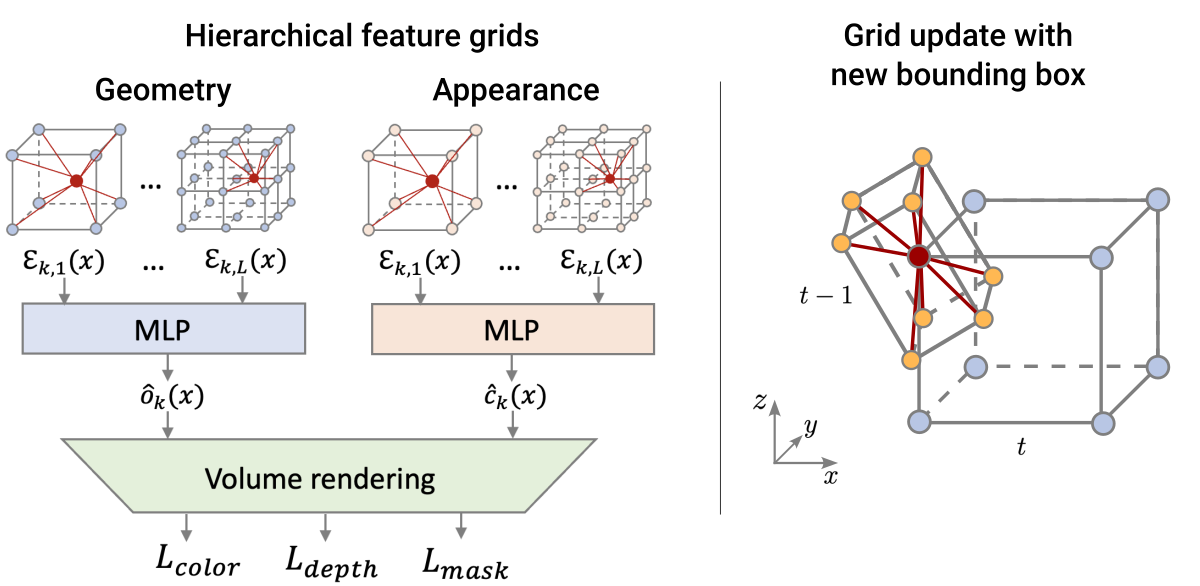}
    \vspace{-0.5em}
    \caption{
    (Left) Our object-centric representation. 
    Given a 3D point $x$ inside the object bounding box, we predict its occupancy and color values via two small feature grids and MLPs. The object model is trained by volume rendering.
    (Right) Given the mapping between object bounding boxes at times $t-1$ and $t$, we retrieve features in the former feature grid to update the new one.
    }
    \label{fig:model}
    \vspace{-1.5em}
\end{figure}

\paragraph{\textbf{Object-centric representation.}}
For any object $O_k$ in the scene, we have its segmentation masks before the current time $t$.
As the object has a compact surface, we can restrict its extent to a bounded volume.
To this end, we store a low-resolution point cloud of $O_k$, denoted as $\mathcal{P}_{k,1:t}$, using object masks from the first time $O_k$ appears in video $V$ to time $t$.
We then compute a bounding box encompassing $\mathcal{P}_{k,1:t}$, the center and orientation of which are used to define the frame attached to $O_k$.
We denote the rigid transformation from the world frame $w$ to the object frame at time $t$ as $S^k_t=[s^k_t R^k_t, T^k_t]$ where $s^k_t$, $R^k_t$ and $T^k_t$ are respectively the box extent, orientation and position for object $O_k$ at time $t$ in the world frame.
The point cloud $\mathcal{P}_{k,1:t}$, bounding box and matrix $S^k_t$ are updated with time as more parts of $O_k$ are seen in the video.

We represent $O_k$ by a function mapping a 3D point in its bounding box to a color and an occupancy value, as illustrated on the left of~\Cref{fig:model}.
Specifically, the function is composed of two separate grid encoders and MLP decoders, one predicting occupancy for the geometry and the other predicting colors for appearance. 
Each grid is a dense multi-resolution grid, similarly to~\cite{muller2022ingp} though not using a hash index, which impedes sequential optimization, and encoding only shape or color.
It fully covers the bounding box of $O_k$, its boundaries being the ones of that bounding box.
The grid stores features at $L$ different levels.
At level $l=1, ..., L$, it has $N_l=N_0\gamma^{l-1}$ vertices per side, with $N_0$ the side size of the coarsest level and $\gamma$ a scale factor, and each vertex stores a single feature scalar $E_{k, l}$ in a table.
When querying the encoding of a 3D point $x$ in the grid, we retrieve the stored features for the vertices of the cell $x$ falls in at each level $l$, trilinearly interpolate them and concatenate the $L$ resulting scalars.
The output embedding $\mathcal{E}_k(x) \in \mathbb{R}^L$ is differentiable with respect to the stored features in $E_{k,l}$, which are optimized variables.
We feed $\mathcal{E}_k(x)$ to an MLP which predicts either an occupancy value $\hat{o}_k(x)$ as in~\cite{oechsle2021unisurf} or a color $\hat{c}_k(x)$. 
Surfaces are defined as the $\frac{1}{2}$-level set of the occupancy function.
Note that, following prior work~\cite{sucar2021imap,zhu2022niceslam,johari2023eslam,kong23vmap,sandstrom2023pointslam}, we do not model the color dependency on the viewing direction.

\vspace{-1.5em}
\paragraph{\textbf{Updating feature grids with new views.}}
As the feature grid for object $O_k$ is closely aligned with the bounding box of the object, it must be adapted to cover the extended bounding box with new views.
\noindent Specifically, given object frames at previous time $S_{t-1}^k$ and the current time $S_{t}^k$, we first compute the mapping between the previous and current bounding box $\Delta S_{k, t-1, t} = (S^k_{t-1})\mbox{}^{-1} S_{t}^k$.
We then map each vertex $v$ of the new grid to the former bounding box using $\Delta S_{k, t-1, t}$ and retrieve the corresponding interpolated feature in the former grid, which we store as the feature for $v$ in the new grid.
The right part of \Cref{fig:model} illustrates the process.
We assign random values to vertices that fall out of the previous bounding box.
In this way, the interpolated feature grid replaces the previously stored one to allow incremental updating.
The optimization continues with new feature grid parameters. 

\vspace{-1.5em}
\paragraph{\textbf{Training objectives with volume rendering.}}
\label{par:volume-rendering}
We employ differentiable volume rendering for training.
First, during the mapping sequence, we store a buffer of $20$ keyframes as well as the two running frames for each object $O_k$ as in vMAP~\cite{kong23vmap}.
At each time $t$, we randomly select a set of frames $\mathcal{F}_{k, t}$ in this buffer, and sample pixels that are part of the masks of $O_k$.
Then, given a camera pose $T$ and a pixel $u$, we sample $N$ increasing depth values on the ray $T\cdot u$, among which $N_{surf}$ values follow a normal distribution centered at the measured depth value $D(u)$ with a manually set variance $\sigma^2$ and the other $N - N_{surf}$ values are uniformly distributed between the bounding box's closest border to the camera and $D(u) - 3\sigma$.
Having a bounding box around each object leads to a more efficient sampling by not querying many points in empty space, unlike vMAP~\cite{kong23vmap}.
These points are then expressed in the object frame and fed to the density function to get occupancy and color values $\hat{o}_{k,i}$ and $\hat{c}_{k,i}$. 
We compute the ray termination weight at the i-th point on a ray as $w_{k, i} = \hat{o}_{k, i} \prod_{j<i} (1 - \hat{o}_{k, j})$ and obtain the pixel color $\hat{C}_k$, depth $\hat{D}_k$, mask $\hat{M}_k$ and depth variance $\hat{V}_k$ of object $O_k$ with classical volume rendering (see formulas in the supplementary material).

We optimize the model with cost functions penalizing the difference between the inputs and the renderings for color $\mathcal{L}_{col}$, depth $\mathcal{L}_{depth}$ and mask $\mathcal{L}_{mask}$ for each pixel $u$ (detailed in the supplementary material):
\vspace{-0.8em}
\begin{equation}
    \mathcal{L} = \frac{1}{Z} \sum_{k=1}^K \sum_{u\in \mathcal{F}_{k, t}} \mathcal{L}_{depth} + \lambda_{col} \mathcal{L}_{col} + \lambda_{mask} \mathcal{L}_{mask}.
    \label{eq:loss-online}
    \vspace{-0.8em}
\end{equation}
where $K$ is the number of objects and $Z=K|\mathcal{F}_{k,t}|$.
Note here that each object is independent from the others in the final loss, allowing their optimizations to be parallelized.

\subsection{Integrating prior object-centric neural shape models}
\label{subsec:init-database}
\vspace{-0.5em}
In order to enhance accuracy and shape completion and benefit from knowledge accumulated during previous mapping sessions, we propose to leverage 3D shape priors for online object reconstruction.
We construct an object library containing a set of $N$ objects $\{M_1, ..., M_N\}$ that have already been mapped, be it during previous mapping episodes or with images coming from 3D meshes.
For each object, we store its neural implicit representation (shape feature grid and MLP), a coarse point cloud, and features for matching.
During new mapping sessions, we first search for objects in the library that may appear in each frame and compute initial poses. 
If matches are found, we then texture and complete the initial shapes if necessary, taking a special care in not losing details seen only during previous episodes.

\begin{figure}[t]
    \centering
    \includegraphics[width=\linewidth]{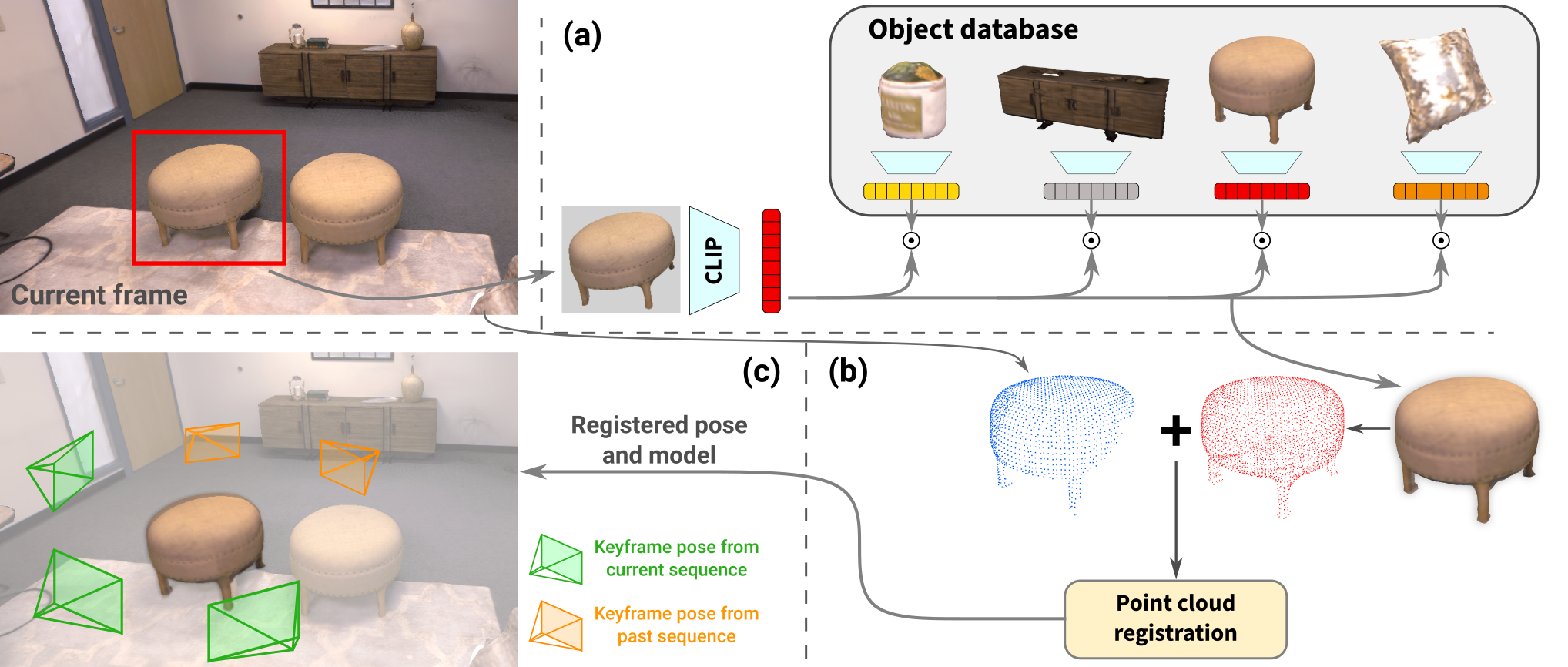}
    \vspace{-2em}
    \caption{\small{
    Overview of the procedure to integrate prior object models.
    {\em (a)} Retrieval: given a newly segmented object, we retrieve the most similar object in the object library via CLIP embedding.
    {\em (b)} Registration: we get an aligned pose of the retrieved object via point cloud registration.
    {\em (c)} Shape refinement: we refine the initial shape model with novel views while additionally synthesizing keyframes from retrieved object models to not lose shape details. 
    }}
    \label{fig:reuse_pipeline}
    \vspace{-1.5em}
\end{figure}

\vspace{-1.2em}
\paragraph{\textbf{Building an object library.}}
We consider two types of resources to obtain objects for our object library.
The first type is from an existing database of 3D meshes, which is for instance relevant in industrial environments where some objects have already been precisely scanned (\eg,~\cite{hodan2018bop}) and their geometry can be reused with no further updating. 
While this database can be of any size, our initialization procedure, which we present next, focuses on reusing the exact same object as the one in the current view, bringing little interest in using large-scale databases with most objects never observed, unlike other works~\cite{runz2020frodo,shan2021ellipsdf,sucar2020nodeslam,wang2021dspslam,zou2022objectfusion}.
We generate synthetic photo-realistic RGB-D images and segmentation masks for each object, taken at multiple random poses, using the BlenderProc renderer~\cite{Denninger2023blenderproc}. 
These images almost fully cover the complete object.
The second type is from previous mapping sessions, \eg, another video stream capturing the same scene, or one with objects shared with the current scene, from different angles. 
This scenario is practical in real world applications where a robot enters a room several times to reconstruct it, not necessarily with a 360° scan, stores object models at the end of each video and then relocates and completes them with novel viewpoints, a setting similar to~\cite{watkins2022mobilemanipulation}. 
Compared to objects in the first type, the objects here are likely to only be partially observed. 
We use the same method described in~\cref{subsec:object-centric} to reconstruct objects from images.
Therefore, each object $M_j$ in the library contains its appearance and shape feature grids and MLPs, as well as its bounding box extent, a low-resolution point cloud which we obtain by rendering and backprojecting depth maps from the object model, and the poses in the object frame of keyframes stored during the model acquisition in previous sequences.
Note that storing and optimizing geometry separately from appearance is possible thanks to their disentanglement into two feature grids and MLPs, as explained above.
We further extract global and local features for each object which are later used for retrieval and registration.
For the retrieval part, we render RGB images with the model from some stored keyframe poses and compute CLIP embeddings~\cite{radford2021clip} for each view. 
We store the averaged embedding of all the views as the single retrieval vector.
For registration, we compute normals in the stored point cloud and extract and store FPFH features~\cite{rusu2009fpfh}.

\vspace{-1.75em}
\paragraph{\textbf{Retrieval and registration.}}
During a new sequence, when a novel object $O_k$ is seen in a frame, we identify the most similar object in our object database and align it with the current view.
First, we crop the current RGB frame around $O_k$, replace the background by some fixed color and compute a CLIP embedding for this image.
We retrieve the $m=3$ objects in our database that have the same semantic category and highest cosine similarity to $O_k$ and filter these matches based on a threshold on their matching scores.
We then register the retrieved point clouds with the backprojected current partial view of the object thanks to Ransac~\cite{fischler1981ransac} followed by a point-to-plane ICP~\cite{chen1992icppoint2plane} relying on normals and likewise filter fitness scores larger than another threshold.
We additionally reproject the registered point clouds in the camera view and ensure that it coincides well with the object mask and that all the points are further away from the camera than the input depth map.
If these conditions are filled, we consider these retrieval and registration to be successful, $M_j$ and $O_k$ being then assumed to be the same object.
We finally initialize $O_k$ as $M_j$, using the registered pose as the object pose in the world frame, and fix the MLPs, and optionally the feature grids, for the rest of the sequence.
If the cosine similarity or fitness thresholds are not reached, we start reconstructing $O_k$ from scratch and retry this initialization when more of the object is visible, \ie, when the bounding box is grown one more time.

\vspace{-1.6em}
\paragraph{\textbf{Synthesizing keyframes on the fly.}}
Initializing an object from a previous model and updating it with solely novel views as explained previously gradually loses the shape information from the original model. 
Yet, storing and reusing all past keyframes is infeasible for long sequences with numerous objects, which is why we only maintain a buffer of keyframes per object.
To alleviate this issue, we instead synthesize on the fly views from the retrieved model and add them to the keyframe buffer.
Inspired by previous work~\cite{po2023instantcontinuallearning}, at each optimization step and for each initialized object, we sample $\frac{|\mathcal{F}_{k,t}|}{2}$ camera poses among those stored in the database and render color, depth and mask from these poses using a fixed copy of the reused model.
These synthesized views are particularly important for object parts not seen in the current video.
Our experiments show that this allows to complete the object with novel views while preserving the original shape details.

\vspace{-0.5em}
\section{Experiments}
\label{sec:experiments}

\vspace{-0.25em}
\subsection{Experimental setting}
\label{exp:setting}

\vspace{-0.4em}
\paragraph{Datasets.} 
We evaluate the proposed approach on two datasets of indoor scenes, Replica~\cite{straub2019replica,sucar2021imap} and ScanNet~\cite{dai2017scannet}, and on our own sequences.
The original Replica dataset consists  of synthetic, noiseless 2000-frame RGB-D videos, one for each of 8 synthetic environments, along with the corresponding camera poses and segmentation masks~\cite{straub2019replica}. 
Ground-truth meshes are also available for quantitative evaluation. 
To get more meaningful evaluations, we manually remove from this dataset ground-truth meshes with fewer than 50 vertices, \eg, product tags, and clean a few noisy meshes.
The ScanNet dataset consists of real RGB-D videos of various durations recorded with a moving tablet. It also comes with camera poses from BundleFusion~\cite{dai2017bundlefusion} as well as manually annotated object masks which however are still noisy. 
Our own sequences are recorded in our lab with a Realsense D435 RGB-D camera filming static scenes of few objects. We compute camera poses with ORB-SLAM2~\cite{murartal2017orbslam2} and extract and track consistent masks with Tracking-Anything~\cite{cheng2023trackinganything,liu2023groundingdino}.

\vspace{-1.25em}
\paragraph{Implementation details.}
For our object models, the feature grid contains $L=3$ levels with $N_0=16$ and $\gamma=1.5$, and the MLPs consist respectively of 1 and 2 64-neuron hidden layers for geometry and color.
As this work focuses on object representations, we use the background model from vMAP~\cite{kong23vmap}, which is a single MLP.
For our object library, we compare two resources to get object models: one using 3D meshes, which are the ground-truth Replica meshes extracted by vMAP~\cite{kong23vmap}; the other using another video sequence acquired with different viewpoints for each environment, which was also generated by vMAP~\cite{kong23vmap}.
We use the ground-truth retrieval and registration in the experiments if not otherwise mentioned.
In optimization, we set $\lambda_{mask}=10$ and take $\lambda_{color}=5$ on Replica and $2$ on ScanNet.
We provide additional details about our implementation in the supplementary material.

\vspace{-1.25em}
\paragraph{Metrics.}
We measure the quality of reconstructions on meshes extracted from the Replica sequences using {\em accuracy} and {\em completion} in {\em cm} as well as {\em completion ratio} with various thresholds, as in prior work~\cite{kong23vmap}.
Denoting the reconstructed and ground-truth meshes as $R$ and $G$, accuracy is defined as the average distance between points in $R$ and their nearest neighbour in $G$ and completion is the average distance between points in $G$ and their nearest neighbour in $R$.
We consider two types of ground-truth meshes in our evaluation.
The first is so-called seen parts where we cull the original mesh by removing vertices that never appear in the input video. We focus on the accuracy metric for the seen parts.
The second is the whole mesh including both seen and unseen parts from the video. 
It is more meaningful to measure the completion metrics with the whole mesh when using object priors that inform about parts unseen in the current video.
In addition, unlike prior works~\cite{sucar2021imap,zhu2022niceslam,kong23vmap,johari2023eslam,sandstrom2023pointslam} that subsample vertices on the ground-truth meshes for evaluation, we consider all the vertices to remove any impact from this sampling.
Computation time is also measured and discussed in the supplementary material.
For the ScanNet dataset, since no ground truth mesh exist and object masks are very noisy, our evaluations are thus only qualitative.

\begin{table}[t]
\scriptsize
\centering
\tabcolsep=0.07cm
\begin{tabular}{c|c|cc|cccc}
\toprule
& \multirow{2}{*}{\begin{tabular}[c]{@{}c@{}}Object\\prior\end{tabular}}
& \multicolumn{2}{c|}{Seen parts} 
& \multicolumn{4}{c}{Whole objects}
\\
&  & Acc.$\downarrow$ & \multicolumn{1}{c|}{Comp.$\downarrow$} & Acc.$\downarrow$ & Comp.$\downarrow$ & CR 1cm$\uparrow$ & CR 5mm$\uparrow$
\\
\midrule
TSDF*~\cite{curless1996tsdf}
& --- & \textbf{0.61} & 0.38 & \textbf{0.59} & 3.07 & 73.7 & 69.2 \\
\midrule
vMAP*~\cite{kong23vmap}
& --- & 1.32 & 0.79 & 2.31 & 2.10 & 73.0 & 52.7 \\
\midrule
\multirow{3}{*}{\begin{tabular}[c]{@{}c@{}c@{}}Ours\end{tabular}}
& ---         & 0.82 & 0.34 & 2.31 & 2.43 & 81.3 & 75.9 \\
& 3D meshes   & 0.78 & \textbf{0.26} & 1.25 & \textbf{0.29} & \textbf{98.7} & \textbf{93.9} \\
& Prior video & 0.81 & 0.33 & 2.28 & 1.36 & 85.2 & 80.1 \\
\bottomrule
\end{tabular}
\vspace{-1em}
\caption{
Object-level reconstruction performance averaged over 8 Replica scenes evaluated with seen parts and whole objects as ground-truth meshes. Our results with object priors rely on ground truth meshes and shape from previously viewed videos. 
CR stands for Completion Ratio.
Methods with * are reproduced results from the official code bases.
}
\label{tab:replica_objects}
\vspace{-2.35em}
\end{table}

\vspace{-1.35em}
\paragraph{Baselines.}
We compare our method with prior object-level reconstruction methods.
We run all the compared methods with their released codes and the provided ground-truth camera poses for fair comparison.
vMAP~\cite{kong23vmap} and RO-MAP~\cite{han2023romap} are the only object-level neural implicit RGB-D methods we are aware of.
However, since RO-MAP did not release codes on the evaluated datasets and details lack in the paper to reproduce faithfully their results, we mainly compare with vMAP in the main paper and put the discussion of RO-MAP in the supplementary material. 
We also evaluate an object-level TSDF~\cite{curless1996tsdf} integration.
Most neural implicit RGB-D methods operate at the level of scenes, \eg~\cite{sucar2021imap,johari2023eslam,sandstrom2023pointslam}: we also compare to them in the supplementary material.
We extract object meshes at a resolution of 5mm for all methods using marching cubes~\cite{lorensen1987marchingcubes}, and disable any mesh post-processing. 
Each method is run with 5 different initializations on each environment, and we present results averaged over all runs.

\begin{figure}[t]
    \centering
    \includegraphics[width=0.955\linewidth]{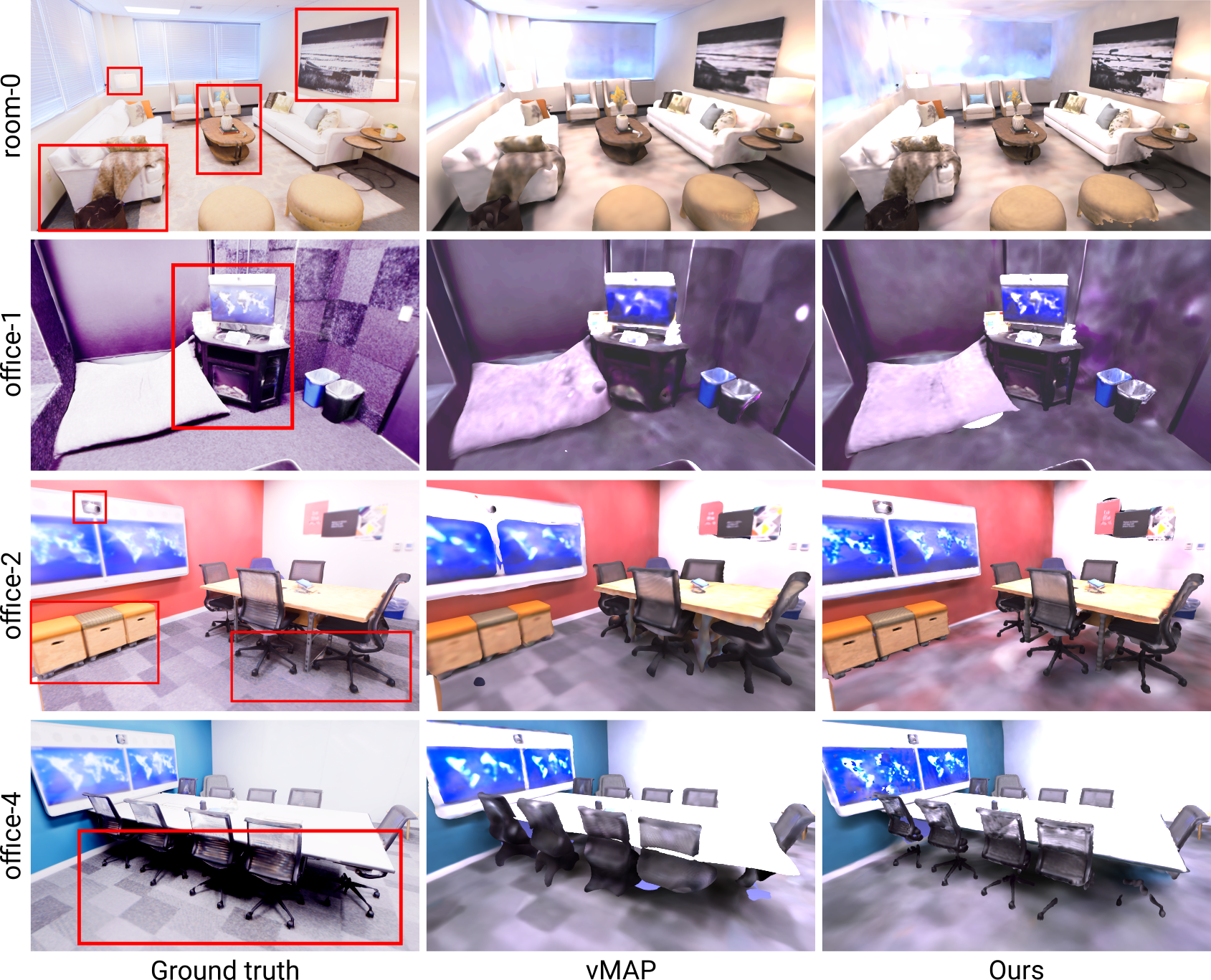}
    \vspace{-0.75em}
    \caption{\small{
    Examples of reconstructions with our method on different Replica scenes, compared to vMAP~\cite{kong23vmap}.
    Our method recovers object geometry that is more faithful to the actual shapes and with better texture.
    }}
    \label{fig:replica-reconstructions}
    \vspace{-1em}
\end{figure}

\begin{figure}[]
    \centering
    \includegraphics[width=0.91\linewidth]{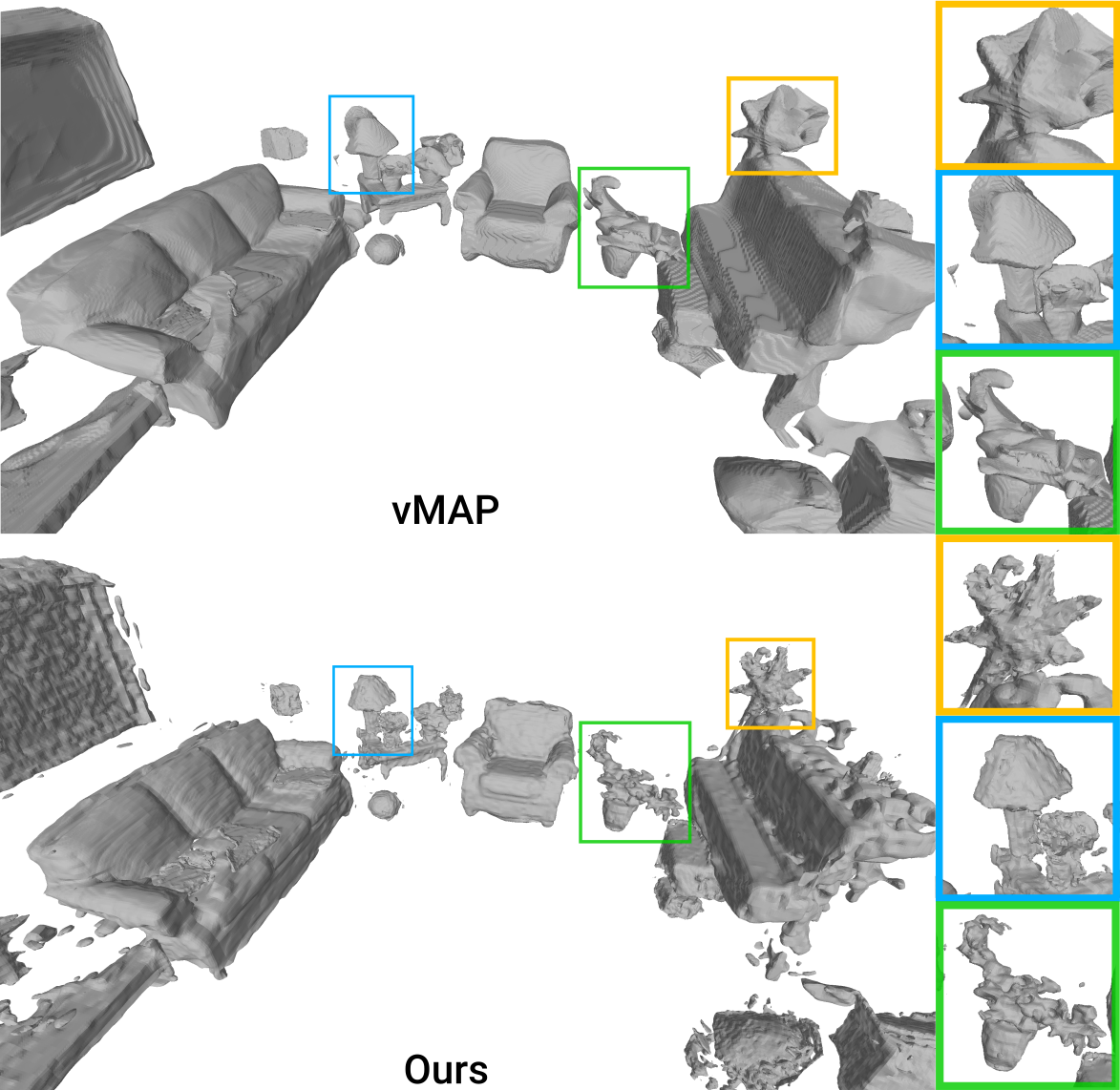}
    \vspace{-0.5em}
    \caption{\small{
    Reconstruction of a ScanNet sequence with vMAP and our method, with close-up views on some parts. 
    Our method recovers more accurate geometries than vMAP, which over-smoothes surfaces
    - see in particular the piano and plant on the right or the sofa on the left - though it is a bit more sensitive to ScanNet's noisy inputs.
    }}
    \label{fig:scannet-reconstructions}
    \vspace{-1.8em}
\end{figure}

\vspace{-0.5em}
\subsection{Comparison with the state of the art}
\label{exp:reconstruction}

\vspace{-0.5em}
\paragraph{\textbf{Replica.}}
\Cref{tab:replica_objects} presents the object-level reconstruction performance of vMAP~\cite{kong23vmap}, TSDF~\cite{curless1996tsdf} and variants of our method.
When no object shape priors are considered in both approaches, our method outperforms vMAP across all metrics for the seen parts, with a relative improvement of around 40\% in accuracy and 60\% in completion. 
This highlights the superior capability of our object-centric neural implicit model in faithfully reconstructing objects from video streams.
For metrics on the whole objects, we observe a decrease in completion for our method. 
This discrepancy arises since vMAP uses pure MLPs to predict smooth occupancy values in space, then extracting regular surfaces beyond the seen parts.
In contrast, while the feature grid we employ is more powerful, it is also more local and only predicts surfaces for the visible parts of objects.
However, completion ratios are much higher with our model than vMAP on whole meshes, with an increase of $23.2\%$ at 5mm, showcasing superior capability to accurately fit seen ground-truth mesh parts. 
Compared to TSDF, our approach is less accurate on seen parts and whole meshes, but is more complete and has higher completion ratios.
As our method relies on MLPs and multi-scale continuous feature grids, it may slightly benefit from some surface regularity beyond seen parts, unlike TSDF, explaining this gap.
Adding object priors mitigates this issue and leads to a substantial improvement in completion metrics for whole object evaluation.
When compared to using full 3D meshes as shape priors, using object models fitted on prior video sequences results in a decrease in performance.
This can be attributed to the fact that the two video sequences for each scene cover very similar viewpoints of objects, resulting in limited potential improvement on these metrics. 
We show a few examples of our reconstructions on scenes of the Replica dataset in~\Cref{fig:replica-reconstructions}.
Our object representations recover geometry that is more accurate and faithful to the original scene.

\begin{figure}[!t]
    \centering
    \includegraphics[width=0.98\linewidth]{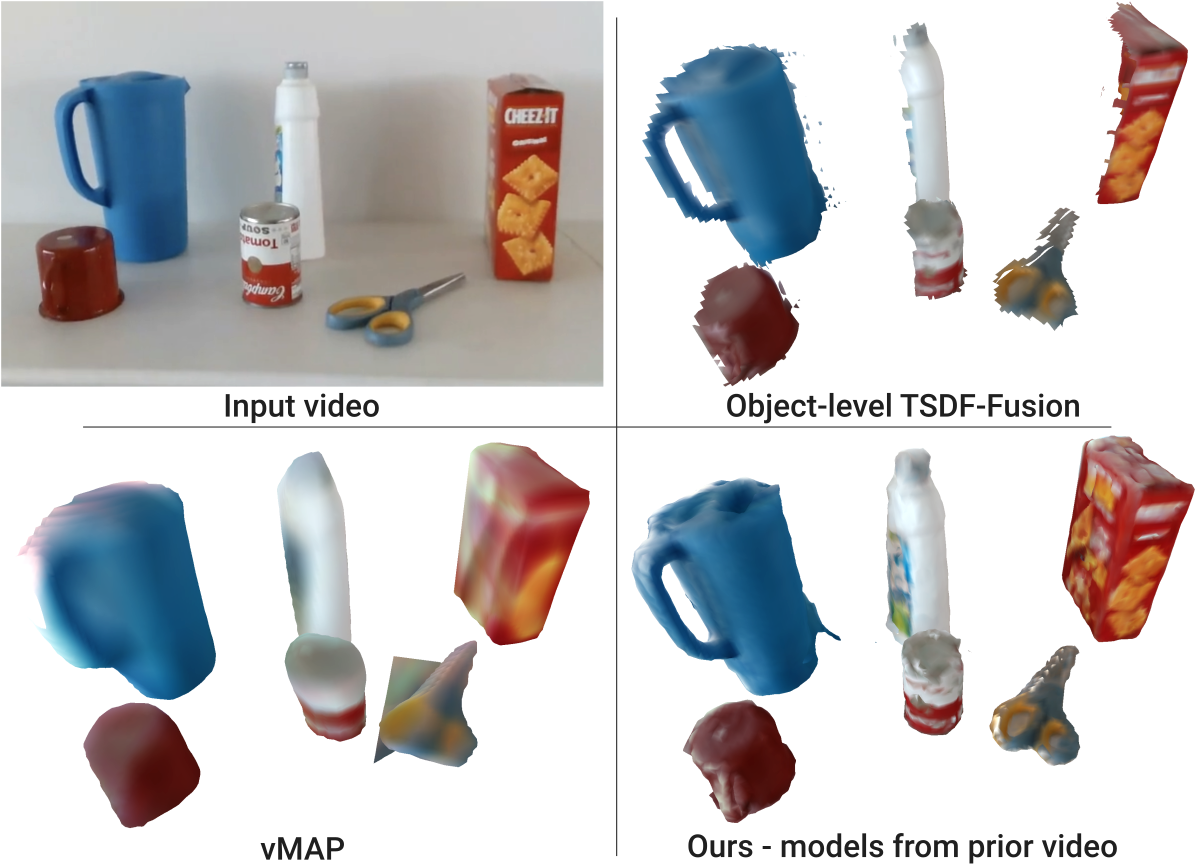}
    \vspace{-0.8em}
    \caption{\small{
    Reconstruction of a self-captured sequence with object-level TSDF, vMAP and our method using a prior video.
    }}
    \label{fig:real_world_self}
    \vspace{-1.8em}
\end{figure}

\vspace{-1.25em}
\paragraph{\textbf{ScanNet.}}
For real-world scenes, we show a reconstructed scene from the ScanNet dataset in~\Cref{fig:scannet-reconstructions}.
Again, our reconstructions are closer to the actual geometry of the scene than vMAP.
However, our approach is more sensitive to noise in the input segmentation masks and depth images (see supplementary material for noisy input data).
Since our bounding boxes are computed from these inputs and grown as soon as a single segmented point for the object falls outside it, an ill-segmented image may have the boxes grow too much and lead to artifacts in areas that are not or little mapped, for which stored features are random.
Our object grid representations also optimize  local features unlike vMAP's MLPs whose weights are shared for a whole object and thus more global.
This implies that our models are optimized locally to fit all points including outliers, whereas vMAP oversmoothes geometries, trading a loss of important object details for a higher robustness to outliers.

\begin{figure}[t]
    \centering
    \includegraphics[width=0.99\linewidth]{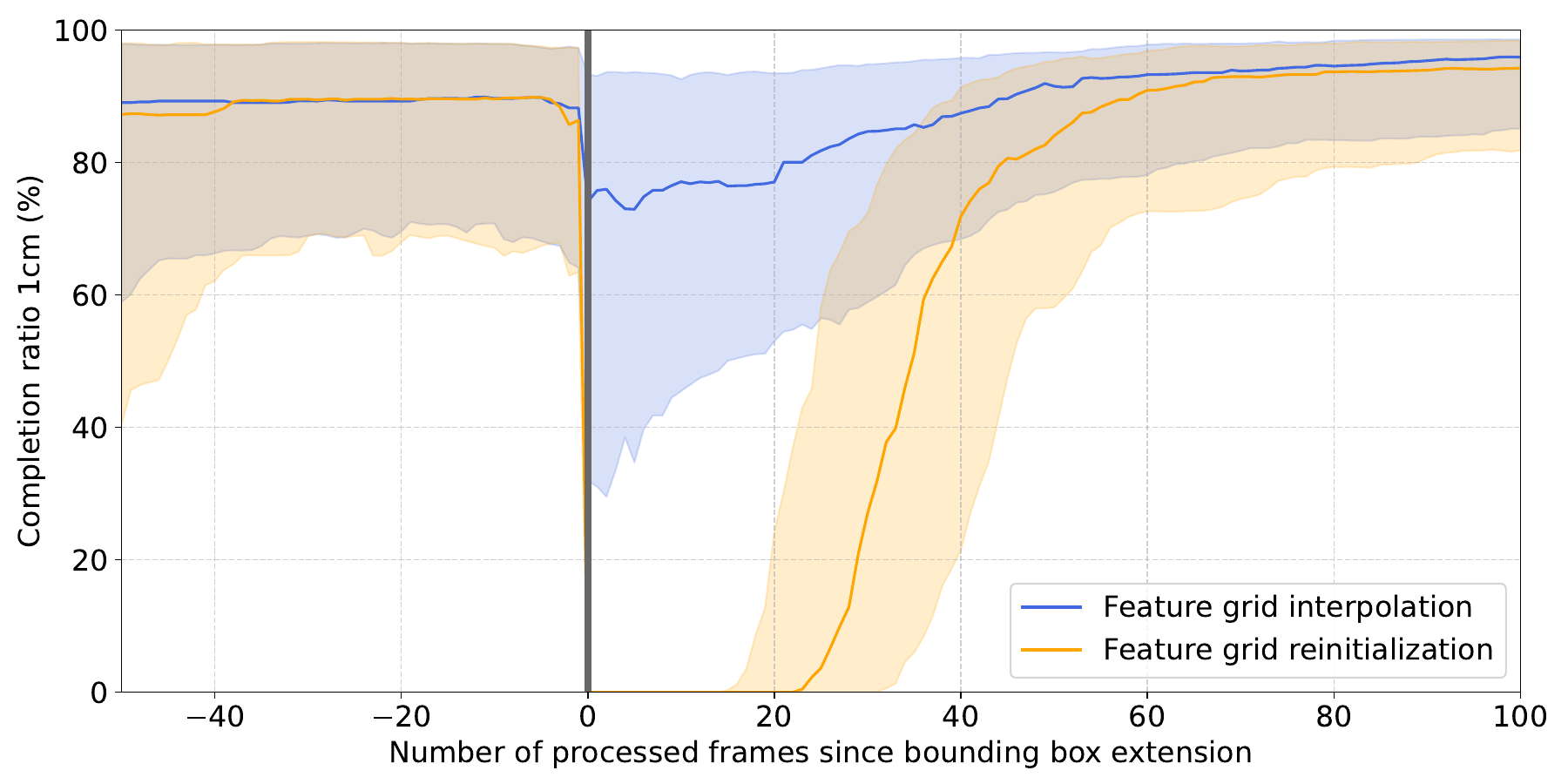}
    \vspace{-0.8em}
    \caption{\small{
    Evolution of the median completion ratio at 1cm before and after object bounding box extensions, either interpolating grid features or reinitializing the grid from scratch.
    Colored areas represent the 20-th and 80-th percentiles.
    Interpolating features leads to much smaller decays and more precise meshes right after box extension (vertical line).
    }}
    \label{fig:grid_interpolation}
    \vspace{-0.8em}
\end{figure}

\begin{figure}[t]
    \centering
    \includegraphics[width=\linewidth]{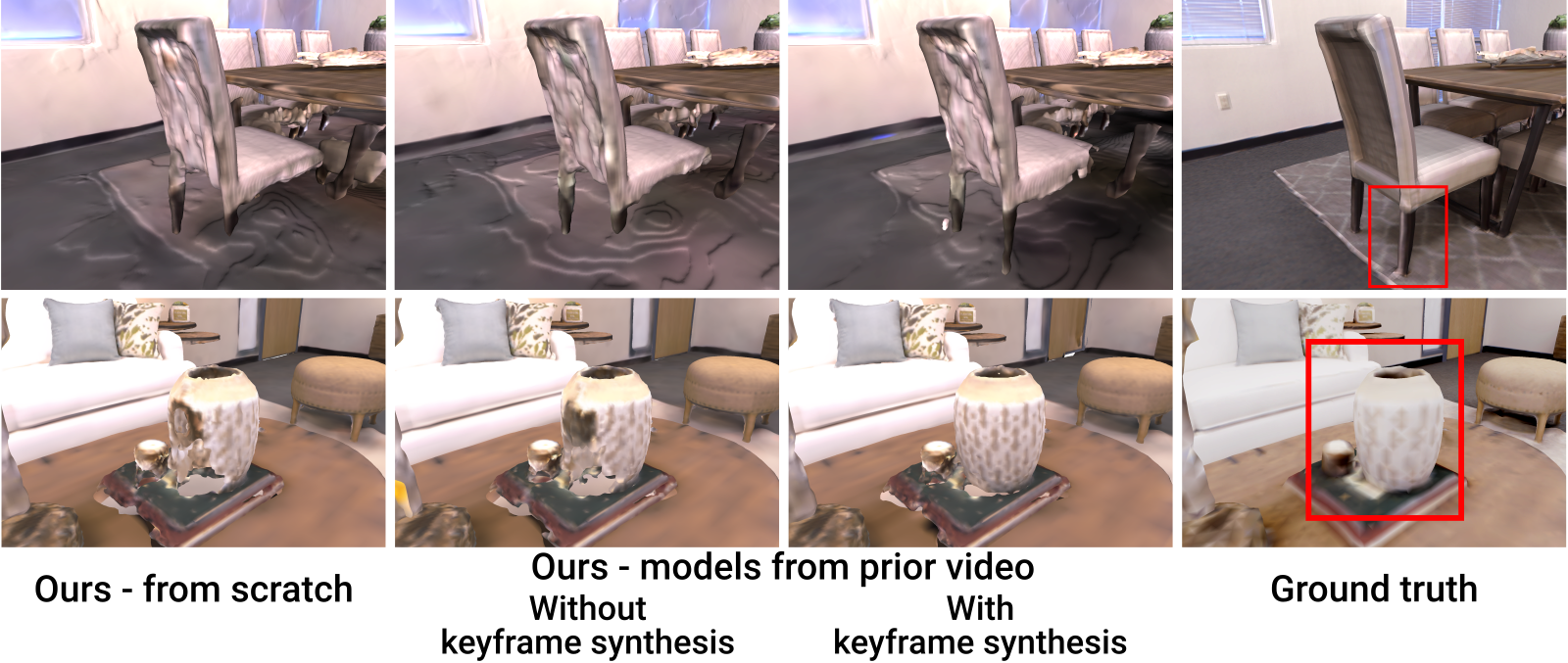}
    \vspace{-1.8em}
    \caption{\small{
    Visualization of completed meshes with our method on Replica scenes.
    Interesting parts are highlighted.
    Note that the back of the chair in the first row is never seen in any video, which explains its poor reconstruction.
    }}
    \label{fig:completion}
    \vspace{-1.8em}
\end{figure}

\vspace{-1.2em}
\paragraph{\textbf{Our own video.}}
We further evaluate the reconstruction of a real-world sequence acquired in our lab, where we compare variants of our model to TSDF~\cite{curless1996tsdf} and vMAP~\cite{kong23vmap}.
The results are shown in~\Cref{fig:real_world_self}.
Again, our base model recovers much finer details than vMAP and has fewer floating artifacts and more regular geometries on parts seen in few frames than TSDF.
When we additionally leverage a library made from another sequence of the same objects captured with different viewpoints, we obtain higher completion for the objects while preserving details for parts that were only seen in the past video.

\vspace{-0.2em}
\subsection{Ablations}
\label{exp:ablations}

\begin{table}[t]
\centering
\tabcolsep=0.09cm
\scriptsize
\begin{tabular}{c|c|cc|cccc}
\toprule
\multirow{2}{*}{\begin{tabular}[c]{@{}l@{}}Object\\ library\end{tabular}}
& \multirow{2}{*}{\begin{tabular}[c]{@{}l@{}}
Keyframe\\synthesis
\end{tabular}}
& \multicolumn{2}{c|}{Seen parts}
& \multicolumn{4}{c}{Whole mesh}
\\
& 
& Acc.$\downarrow$ & \multicolumn{1}{c|}{Comp.$\downarrow$}
& Acc.$\downarrow$ & Comp.$\downarrow$ & CR 1cm$\uparrow$ & CR 5mm$\uparrow$
\\
\midrule
\multirow{3}{*}{\begin{tabular}[c]{@{}c@{}}3D\\ meshes\end{tabular}}
& Fixed models
& 0.77 & \textbf{0.23} & \textbf{1.21} & \textbf{0.26} & \textbf{99.3} & \textbf{97.1} \\
& \xmark & \textbf{0.76} & 0.34 & 1.96 & 1.61 & 82.8 & 76.4 \\
& \cmark & 0.78 & 0.26 & 1.25 & 0.29 & 98.7 & 93.9 \\
\midrule
\multirow{2}{*}{\begin{tabular}[c]{@{}c@{}}Prior\\ video\end{tabular}} 
& \xmark & 0.80 & 0.34 & 2.16 & 1.82 & 82.5 & 77.0 \\
& \cmark & 0.81 & 0.33 & 2.28 & 1.36 & 85.2 & 80.1 \\

\bottomrule
\end{tabular}
\vspace{-0.75em}
\caption{\small{
Ablations of the keyframe synthesis part of our method, 
using different object libraries.
Metrics are computed at the level of objects and retrieval and registration are ground truth.
In the first row, the whole object models are frozen.
}}
\label{tab:replica_ablations_forgetting}
\vspace{-2.3em}
\end{table}

\vspace{-0.35em}
\paragraph{\textbf{Feature grid interpolation.}}
We compare two strategies for the grid update at each bounding box extension: reinitializing the grid features~\cite{han2023romap} or interpolating them, as proposed above.
For that, we compute completion ratios at 1cm on seen parts of objects after each frame of each Replica sequence.
We then select the metrics in a window of few frames before and after each box extension and average them over all the objects and their box update steps.
The result is shown on~\Cref{fig:grid_interpolation}.
We observe that reinitializing features requires 20 more frames after the update before extracting a surface from the representation.
Conversely, interpolating features results in a small decay of the metric immediately after the interpolation step, a value that is otherwise reached after fitting for 40 more frames when reinitializing the features.
This shows the importance of our interpolation strategy to get an accurate model at all time.

\vspace{-1em}
\paragraph{\textbf{Fitting with synthesized keyframes.}}
We evaluate in~\Cref{tab:replica_ablations_forgetting} the effectiveness of the proposed strategy to avoid losing original shape details when utilizing the shape prior. 
In the upper block of the table, as the reused object models are obtained from full 3D meshes, simply freezing them achieves high performance in both accuracy and completion.
However, continuing to update them with no further change, as indicated in the second row, 
exhibits a significant drop in completion especially on the whole mesh. 
This underlines the severity of the loss of detail problem.
The proposed strategy plays a crucial role in addressing it for unseen parts, maintaining comparable performance to the unoptimized models, see the third row.
When employing object models fitted on prior video sequences in the bottom block, updating the models, imperfect due to partial observations, becomes necessary. 
In this case, our proposed strategy preserves similar performance on seen parts without compromising completion metric on the whole mesh.
Our strategy thus allows to complete partially seen objects while not deteriorating fully mapped ones. 
As determining whether an object has already been fully seen can be difficult, we demonstrate here that all objects can be fitted within a single framework.
\Cref{fig:completion} shows examples of using object priors from video sequences with and without keyframe synthesis.

\begin{table}[t]
\centering
\scriptsize
\tabcolsep=0.1cm
\begin{tabular}{c|c|cc|cccc}
\toprule
\multirow{2}{*}{\begin{tabular}[c]{@{}c@{}}Object\\ library\end{tabular}}
& \multirow{2}{*}{\begin{tabular}[c]{@{}c@{}}GT retr.\\ \& reg.\end{tabular}}
& \multicolumn{2}{c|}{Seen parts}
& \multicolumn{4}{c}{Whole mesh}
\\
& 
& Acc.$\downarrow$ & \multicolumn{1}{c|}{Comp.$\downarrow$}
& Acc.$\downarrow$ & Comp.$\downarrow$ & CR 1cm$\uparrow$ & CR 5mm$\uparrow$
\\
\midrule
--- & \xmark & 0.82 & 0.34 & 2.31 & 2.43 & 81.3 & 75.9 \\
\midrule
\multirow{2}{*}{\begin{tabular}[c]{@{}c@{}}3D\\ meshes\end{tabular}}
& \cmark & \textbf{0.78} & \textbf{0.26} & \textbf{1.25} & \textbf{0.29} & \textbf{98.7} & \textbf{93.9} \\
& \xmark & 0.84 & 0.45 & 1.97 & 1.94 & 86.4 & 79.9 \\
\midrule
\multirow{2}{*}{\begin{tabular}[c]{@{}c@{}}Prior\\video\end{tabular}} 
& \cmark & 0.81 & 0.33 & 2.28 & 1.36 & 85.2 & 80.1 \\
& \xmark & 0.85 & 0.88 & 2.34 & 2.77 & 81.3 & 75.3 \\

\bottomrule
\end{tabular}
\vspace{-0.75em}
\caption{\small{
Ablations of the retrieval and registration parts of our method, using different object libraries.
Metrics are computed at the level of objects.
The first row is the reconstruction from scratch.
CR stands for Completion Ratio.
}}
\label{tab:replica_ablations_retrieval}
\vspace{-2.4em}
\end{table}

\vspace{-1em}
\paragraph{\textbf{Object retrieval and registration.}}
\Cref{tab:replica_ablations_retrieval} assesses the influence of object retrieval and registration.
Though utilizing ground-truth retrieval and registration enhances performance compared to the method without any object priors, the automatic retrieval and registration method notably degrades overall performance, though leading to more accurate and complete whole meshes when using 3D meshes priors.
In detail, objects are correctly retrieved in 57\% and 78\% for the full 3D mesh and previous video setups respectively, and 22\% and 51\% of objects are both accurately retrieved and registered. 
When retrieval or registration fails, our models are then fitted from scratch, similarly to the first row.
A few large objects are initialized with a wrong pose with our retrieval and registration, explaining the decrease in completion from the last row.
Future work improving automatic retrieval and registration will benefit our method.

\vspace{-1.8em}
\section{Conclusion}
\label{sec:conclusion}
\vspace{-0.5em}
We present an online method to reconstruct scenes at the level of objects from RGB-D video sequences.
Leveraging object masks and camera poses obtained from the video, we adapt neural implicit object models consisting of grid-based occupancy and color fields, which can be incrementally updated with new views of the object.
To further improve reconstruction accuracy and completeness, we build object libraries from prior sequences or available meshes and propose a way to utilize them as shape prior and update the pre-trained object models without forgetting. 
Experiments on Replica and ScanNet datasets as well as real-world sequences demonstrate the effectiveness of our approach.
Limitations of our work and future directions are discussed in the supplementary material.

\vspace{-0.65em}
\section*{Acknowledgment}
\vspace{-0.65em}
This work was performed using HPC resources from GENCI-IDRIS (Grants 2021-AD011012725R1/R2).
It was funded in part by the French government under management of Agence Nationale de la Recherche as part of the “Investissements d’avenir” program, reference ANR-19-P3IA-0001 (PRAIRIE 3IA Institute). JP was supported in part by the Louis Vuitton/ENS chair
in artificial intelligence and a Global Distinguished Professorship at the Courant Institute of Mathematical Sciences
and the Center for Data Science at New York University.

{
    \small
    \bibliographystyle{ieeenat_fullname}
    \bibliography{_main}
}

\clearpage
\renewcommand{\thefigure}{S\arabic{figure}}
\setcounter{figure}{0}
\renewcommand{\thetable}{S\arabic{table}}
\setcounter{table}{0}
\renewcommand{\thesection}{S\arabic{section}}
\setcounter{section}{0}
\setcounter{page}{1}
\maketitlesupplementary

The supplementary material is organized as follows:
we discuss limitations of our method and possible directions to explore in~\cref{sec:limitations}, provide additional comparisons with state-of-the-art methods both quantitatively and qualitatively in~\cref{sec:sota_cmpr} and introduce more implementation details in~\cref{sec:implementation_details}.

\section{Limitations}
\label{sec:limitations}

We have identified three main limitations of our method.
First, our models are highly sensitive to the quality of input masks, which define object bounding boxes extension and may severely worsen reconstructions. 
Correcting masks online from multiple views and the fitted models should robustify the whole model.
Second, our retrieval and registration often struggle when seeing objects from too different viewpoints from those used for the database models. 
Further research on this part, \eg using stronger image feature models for efficient retrieval~\cite{oquab2024dinov2,zhai2023siglip} and 6D object pose estimation for registration~\cite{ornek2023foundpose,wen2024foundationpose}, should allow for more systematic object reuse and reduce computations.
Third, our method focuses on mapping objects given camera poses provided by an external SLAM system without correcting them. 
Extending it to a complete object-level SLAM is an interesting future direction.

\section{Additional comparisons with the state of the art}
\label{sec:sota_cmpr}

\begin{figure*}[t]
    \centering
    \includegraphics[width=0.945\linewidth]{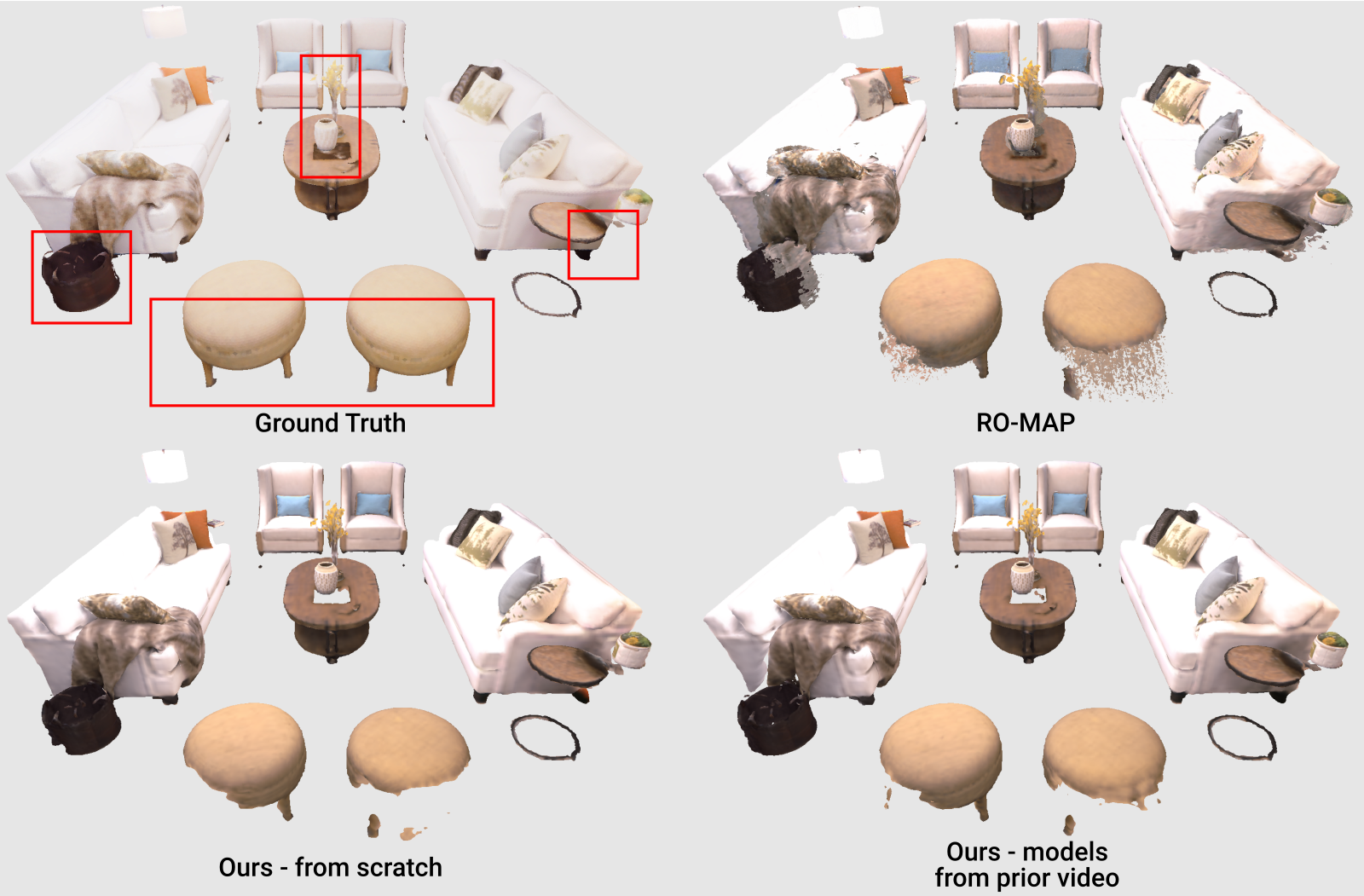}
    \vspace{0.5em}
    \includegraphics[width=0.945\linewidth]{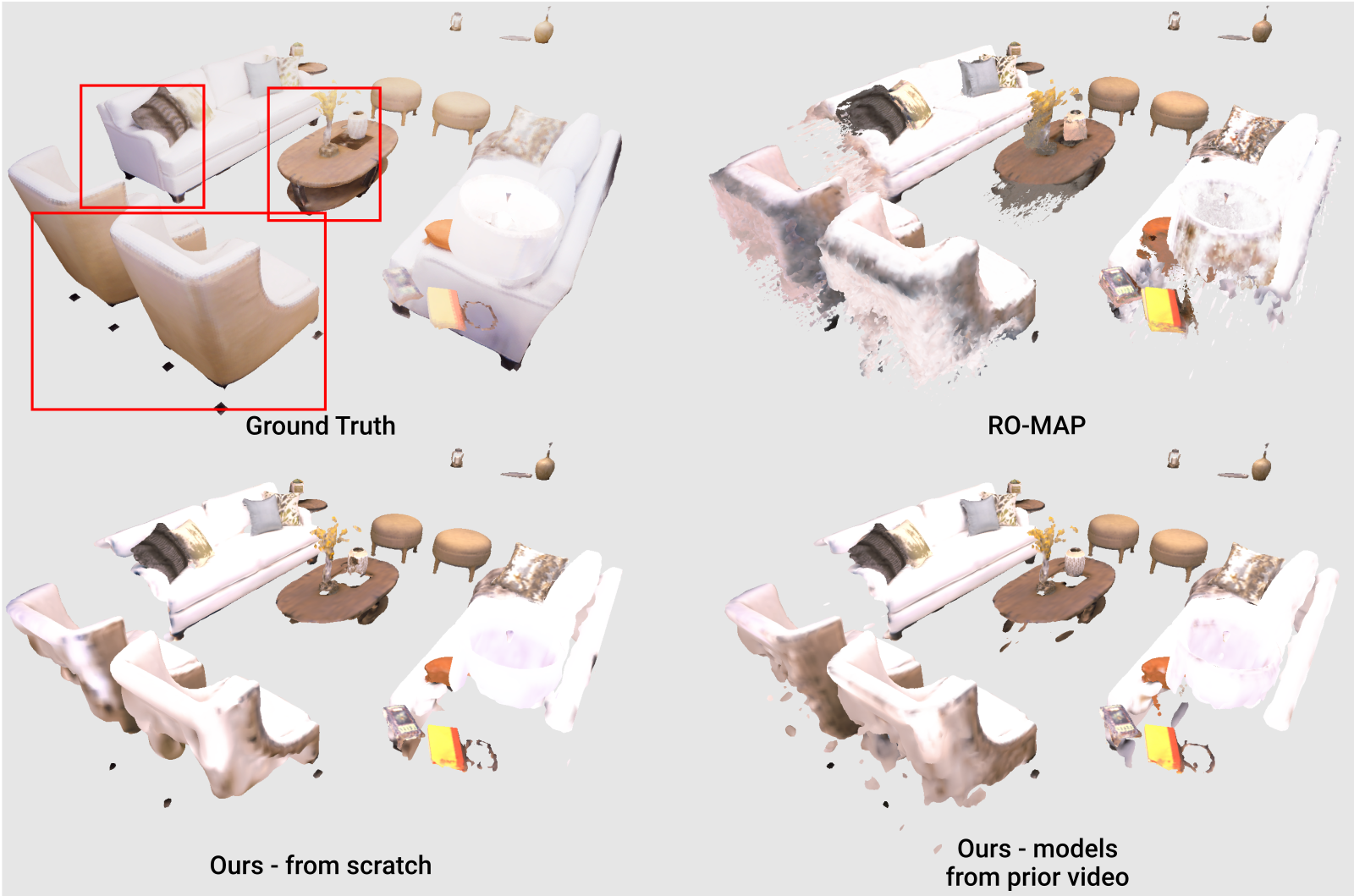}
    \vspace{-1.25em}
    \caption{\small{
    Comparison of object meshes from Replica~\cite{straub2019replica} obtained by RO-MAP~\cite{han2023romap} and our method, using two different viewpoints.
    Meshes for RO-MAP were provided by the authors.
    Our meshes are extracted using marching cubes~\cite{lorensen1987marchingcubes} on a grid of size $64^3$ following RO-MAP's methodology and are restricted to objects reconstructed by RO-MAP.
    Regions of interest are highlighted in red.
    }}
    \label{fig:romap_scene}
\end{figure*}

\begin{figure*}[t]
    \centering
    \includegraphics[width=\linewidth]{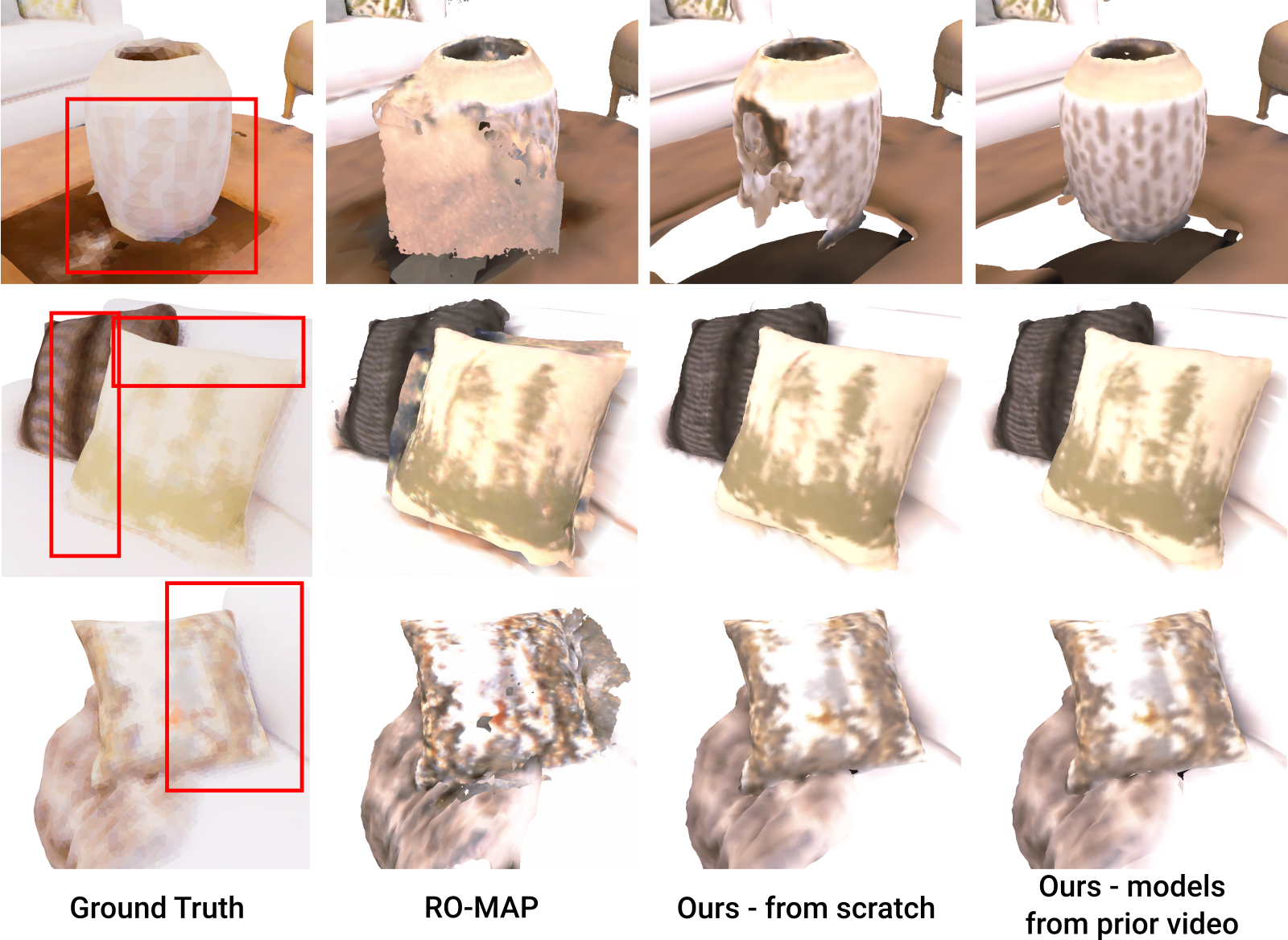}
    \caption{\small{
    Comparison of object meshes obtained by RO-MAP~\cite{han2023romap} and our method.
    Meshes for RO-MAP were provided by the authors.
    Our meshes are extracted using marching cubes~\cite{lorensen1987marchingcubes} on a grid of size $64^3$ following RO-MAP's methodology.
    Regions of interest are highlighted in red.
    Note that the back of the vase on the first row is never seen in the first input sequence and only mapped in the second video which we use for our model shown in the last column.
    }}
    \label{fig:romap_objects}
\end{figure*}

\paragraph{\textbf{Comparison to RO-MAP.}}
We evaluate here our method in the setting of RO-MAP~\cite{han2023romap}.
While the RO-MAP~\cite{han2023romap} paper does not provide all the details about their evaluations and no code has been released to reproduce results on the Replica dataset, we reproduce the closest setting to their work for fair comparison, following discussion with RO-MAP authors.
In this way, we extract meshes for each object using marching cubes~\cite{lorensen1987marchingcubes} on a grid of size $64^3$ as detailed in their paper.
We evaluate two scenes with the same objects, \ie, 23 objects for the scene {\em room-0} and 14 for {\em office-1} as shared by RO-MAP authors.
We present our quantitative results in \Cref{tab:romap_objects} and qualitative examples of a scene and close-ups on few objects in \cref{fig:romap_scene} and \cref{fig:romap_objects}.
The meshes presented here for RO-MAP were shared by the authors.
Our evaluations show that our models computed without any object prior are more accurate and have a better completion ratio at 1cm than RO-MAP, though the metric of completion distance is not so good.
However, as shown in \cref{fig:romap_scene}, RO-MAP objects possess numerous artefacts due to their uniform sampling of points along rays and different losses used, which explains their high accuracy error, very low completion distance and high completion ratio at 5cm, though large parts of objects are unseen in the input videos.
In contrast, our object models are sampled only close to the surface during fitting, leading to much fewer outliers.
Our method also extracts more faithful geometry on contours of objects as shown on \cref{fig:romap_objects}.
Leveraging object priors significantly improves the completion of objects in our method.
While RO-MAP accumulates numerous views before computing a bounding box and restarts optimization from scratch each time these boxes change, our interpolation strategy allows us to reconstruct objects as soon as we observe them, even very partially, and then update their models at each frame without any reinitialization.

\begin{table}[t]
\scriptsize
\centering
\tabcolsep=0.095cm
\begin{tabular}{c|c|c|cccc}
\toprule
& \multirow{2}{*}{\begin{tabular}[c]{@{}c@{}}Object\\prior\end{tabular}}
& \multirow{2}{*}{\begin{tabular}[c]{@{}c@{}}Scene\end{tabular}}
& \multicolumn{4}{c}{Whole objects}
\\
& & & Acc. $\downarrow$ & Comp. $\downarrow$ & CR 1cm $\uparrow$ & CR 5cm $\uparrow$
\\
\midrule
\multirow{2}{*}{\begin{tabular}[c]{@{}c@{}c@{}}RO-MAP${}^{\dag}$\cite{han2023romap}\end{tabular}}
& \multirow{2}{*}{\begin{tabular}[c]{@{}c@{}c@{}}---\end{tabular}}
& room-0 & 3.65 & 0.93 & 69.3 & 98.5 \\
& 
& office-1 & 3.74 & 1.15 & 67.9 & 97.7 \\
\midrule
\multirow{6}{*}{\begin{tabular}[c]{@{}c@{}c@{}}Ours\end{tabular}}
& \multirow{2}{*}{\begin{tabular}[c]{@{}c@{}c@{}}---\end{tabular}}
& room-0 & 2.04 & 2.02 & 73.8 & 90.0 \\
& 
& office-1 & 2.34 & 1.32 & 74.6 & 95.3 \\
& \multirow{2}{*}{\begin{tabular}[c]{@{}c@{}c@{}}3D meshes\end{tabular}}
& room-0 & 1.31 & 0.58 & 86.2 & 99.9 \\
& 
& office-1 & 1.27 & 0.57 & 86.0 & 100.0 \\
& \multirow{2}{*}{\begin{tabular}[c]{@{}c@{}c@{}}Prior video\end{tabular}}
& room-0 & 2.09 & 1.75 & 74.9 & 91.2 \\
& 
& office-1 & 2.43 & 1.11 & 76.4 & 96.3 \\
\bottomrule
\end{tabular}
\caption{
Object-level reconstruction performance compared to RO-MAP~\cite{han2023romap}, using the scenes and settings of RO-MAP's evaluations.
Our results with object priors rely on full 3D meshes and shapes from previously viewed videos. 
Retrieval and registration in these cases are ground truth.
CR stands for Completion Ratio.
Results for RO-MAP are taken from the original paper.
}
\label{tab:romap_objects}
\end{table}

\paragraph{\textbf{Comparison to vMAP.}}
\begin{table}[t]
\scriptsize
\centering
\tabcolsep=0.12cm
\begin{tabular}{c|c|cccc}
\toprule
& \multirow{2}{*}{\begin{tabular}[c]{@{}c@{}}Object\\prior\end{tabular}}
& \multicolumn{4}{c}{Whole objects}
\\
& & \multicolumn{1}{c}{Acc. $\downarrow$} & \multicolumn{1}{c}{Comp. $\downarrow$}
& \multicolumn{1}{c}{CR 1cm $\uparrow$} & CR 5cm $\uparrow$
\\
\midrule
vMAP${}^{\dag}$\cite{kong23vmap}
& --- & 2.23 & 1.44 & 69.2 & 94.6 \\
vMAP${}^*$\cite{kong23vmap}
& --- & 1.84 & 2.32 & 63.6 & 91.5 \\
\midrule
\multirow{3}{*}{\begin{tabular}[c]{@{}c@{}c@{}}Ours\end{tabular}}
& --- & 1.52 & 2.58 & 73.9 & 91.0 \\
& 3D meshes & 1.00 & 0.61 & 86.4 & 99.7 \\
& Prior video & 1.54 & 1.62 & 77.1 & 93.3 \\
\bottomrule
\end{tabular}
\caption{
Object-level reconstruction performance compared to vMAP~\cite{kong23vmap}, reproducing as closely as possible the evaluation setting of vMAP.
Our results with object priors rely on full 3D meshes and shapes from previously viewed videos. 
Retrieval and registration in these cases are ground truth.
CR stands for Completion Ratio.
Results with ${}^{\dag}$ are taken from the original paper and with ${}^*$ are reproduced.
The validity of the difference between the published and reproduced results for vMAP was confirmed by the authors of vMAP.
}
\label{tab:vmap_comparison}
\end{table}

\begin{figure*}[t]
    \centering
    \includegraphics[width=\linewidth]{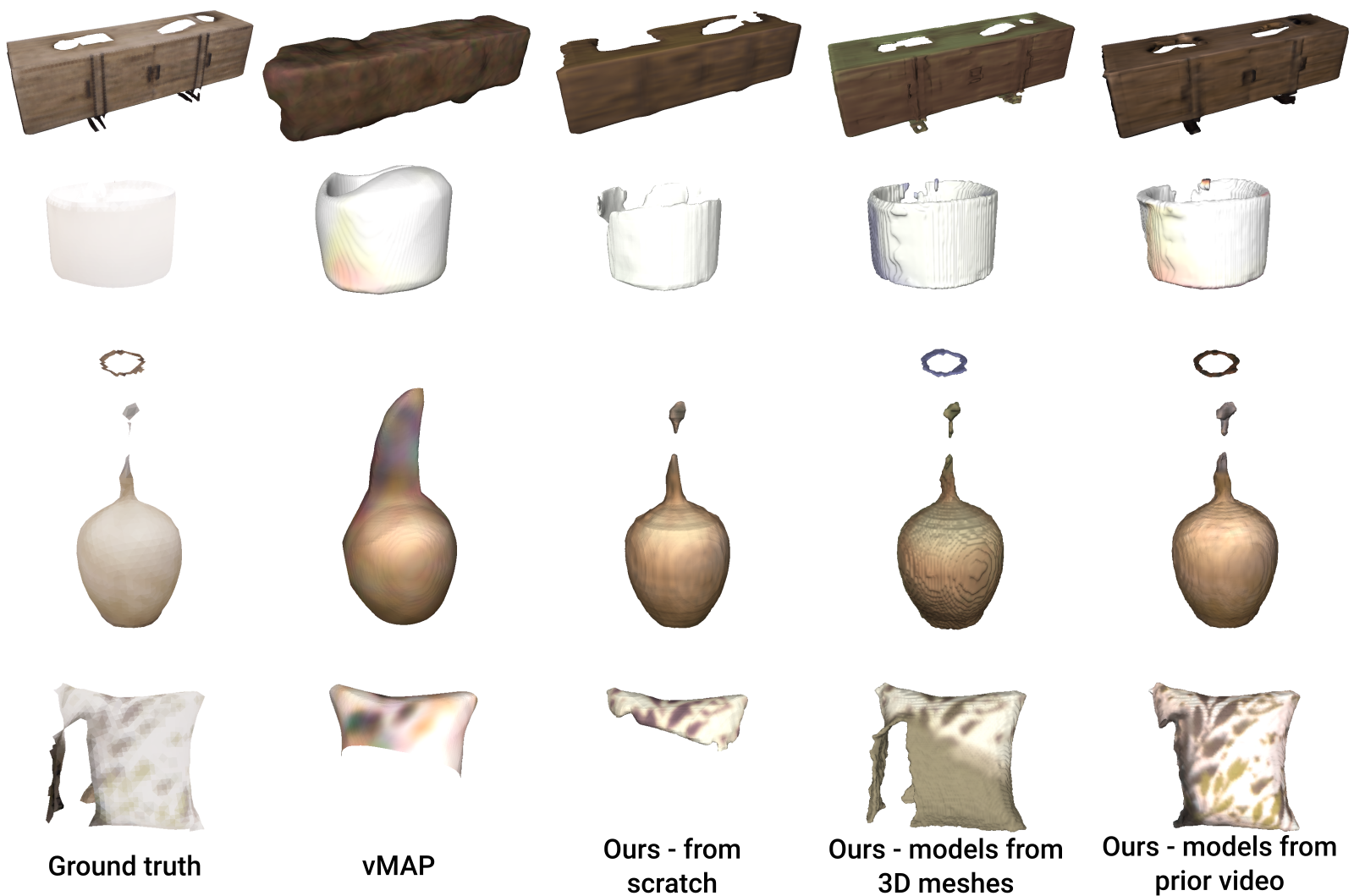}
    \caption{
    \small{
    Comparison of meshes between vMAP~\cite{kong23vmap} and the variants of our method on objects from the Replica dataset~\cite{straub2019replica} after only 50 frames of optimization.
    Meshes are extracted with a resolution of 5mm and our models using object priors rely on ground truth retrieval and registration.
    While our reconstructions from scratch are more accurate than vMAP, adding object priors helps recovering faster the geometry and, optionally, texture of an object.
    }
    }
    \label{fig:early_objects}
\end{figure*}

We further compare our method with vMAP on the same objects, metrics and evaluation procedure as used by vMAP~\cite{kong23vmap}, to provide fair comparison to their published results and confirm that our updated setup proposed in the main paper does not artificially improve our performances over other baselines.
Differing from the evaluations of the main paper, we consider here all objects in each scene, regardless of their size or presence of noise.
We still extract object meshes at a 5mm resolution but set a maximum of 256 points per size and subsample 10k points in the GT and reconstructed meshes to compute metrics instead of considering full meshes.
Results are presented in \cref{tab:vmap_comparison}.
For vMAP, we provide results taken from the original paper as well as reproduced results from the released code.
Though these reproduced results differ from the published ones, we confirmed the validity of this difference with the main authors of vMAP.
Reconstruction results with our models also confirm the trend observed in the main paper with our other evaluation setup: our models optimized from scratch are again more accurate than vMAP and have a higher completion ratio at 1cm.
Their performances are similarly further increased by leveraging object shape priors.
We finally present in~\cref{fig:early_objects} few visualizations of object meshes extracted after only 50 frames processed in the input sequence.
In that setting, vMAP~\cite{kong23vmap} tends to produce oversmoothed meshes which lack geometric and appearance details, \eg, the texture of wood on the first row or the cushion's colored leaves on the last row.
Conversely, our model from scratch recovers more faithful geometry and early texture for these objects.
Adding object shape and texture prior further boosts the representations, leading to more complete and better textured models, in particular for our models obtained from a previous video.

\begin{table}[t]
\centering
\footnotesize
\setlength{\tabcolsep}{0.25em}
\resizebox{\columnwidth}{!}{%

\begin{tabular}{c|c|c|cc|cccc}
\toprule
&  & \multirow{2}{*}{\begin{tabular}[c]{@{}c@{}}Object\\prior\end{tabular}} & \multicolumn{2}{c|}{Seen parts}
& \multicolumn{4}{c}{Whole mesh}
\\
& Objects & 
& Acc. $\downarrow$ & Comp. $\downarrow$
& Acc. $\downarrow$ & Comp. $\downarrow$ & CR 5cm $\uparrow$ & CR 1cm $\uparrow$
\\
\midrule
TSDF*~\cite{curless1996tsdf}
& \xmark
& --- 
& 0.55 & 0.41 & 2.66 & 4.31 & 87.0 & 81.8
\\
iMAP*~\cite{sucar2021imap}
& \xmark
& --- 
& 0.92 & 0.91 & 1.89 & 2.42 & 90.1 & 74.7
\\
ESLAM*~\cite{johari2023eslam}
& \xmark
& --- 
& 0.69 & 0.59 & 0.71 & 4.33 & 86.1 & 76.3
\\
Point-SLAM*~\cite{sandstrom2023pointslam}
& \xmark
& --- 
& 0.67 & 0.59 & 0.67 & 4.76 & 85.3 & 79.5
\\
\midrule
vMAP*~\cite{kong23vmap}
& \cmark
& --- 
& 1.03 & 0.91 & 2.85 & 2.49 & 91.2 & 75.0
\\
\midrule
\multirow{3}{*}{\begin{tabular}[c]{@{}c@{}c@{}}Ours\end{tabular}}
& \cmark
& --- 
& 0.81 & 0.72 & 2.96 & 2.39 & 91.5 & 78.3
\\
& \cmark
& 3D meshes
& 0.82 & 0.70 & 2.08 & 1.71 & 95.3 & 86.3
\\
& \cmark
& Prior video
& 0.80 & 0.70 & 2.69 & 2.25 & 92.3 & 79.3
\\
\bottomrule
\end{tabular}
}
\vspace{-1em}
\caption{
Averaged scene-level reconstruction metrics on the Replica dataset \cite{straub2019replica}, focusing on the ground truth mesh parts observable in the input sequence (left) and whole scene (right).
Our results with object priors rely on full 3D meshes and shapes from previously viewed videos.
CR stands for Completion Ratio.
Methods with * are reproduced results from the official code bases.
}
\label{tab:replica_scene}
\vspace{-1.5em}
\end{table}

\paragraph{\textbf{Scene-level comparisons.}}
For scene-level comparisons, our baselines are TSDF~\cite{curless1996tsdf} (or rather its reimplementation~\cite{zeng20173dmatch}) with a grid resolution of 1cm, vMAP~\cite{kong23vmap} and its reimplementation of iMAP~\cite{sucar2021imap}, ESLAM~\cite{johari2023eslam} and Point-SLAM~\cite{sandstrom2023pointslam}. 
Again, we run all the compared methods with their released codes and the provided ground-truth camera poses for fair comparison. 
We extract scene meshes at a resolution of 1cm for all methods using marching cubes~\cite{lorensen1987marchingcubes}, and disable any mesh post-processing. 
Each method is run with 5 different initializations on each environment, and we present results averaged over all runs.

Results on the Replica dataset are presented in~\Cref{tab:replica_scene}.
As we reuse vMAP's background model, our method outperforms vMAP on the scene-level as well, aligning with the observed trend in object-level evaluation.
Despite the improvement we achieve in object-level methods, our best object-level method with separate models per object still lags behind the best scene-level methods that consider scenes as a single entity when considering metrics on the seen parts or accuracy on whole meshes.
Since the scene metrics depend heavily on the background quality, our background model which oversmoothes surfaces does not capture well details and explains this gap.
Using a more accurate background model would benefit our method.
However, our models present better completion distance and ratios on whole meshes than scene-level baselines, showing the interest of decomposing scenes in objects and reusing priors.
In addition, it is worth noting that all the neural implicit model-based approaches perform worse on seen parts than the traditional TSDF~\cite{curless1996tsdf}, although they showcase advantages on the whole mesh.

\begin{figure*}[t]
    \centering
    \includegraphics[width=0.85\linewidth]{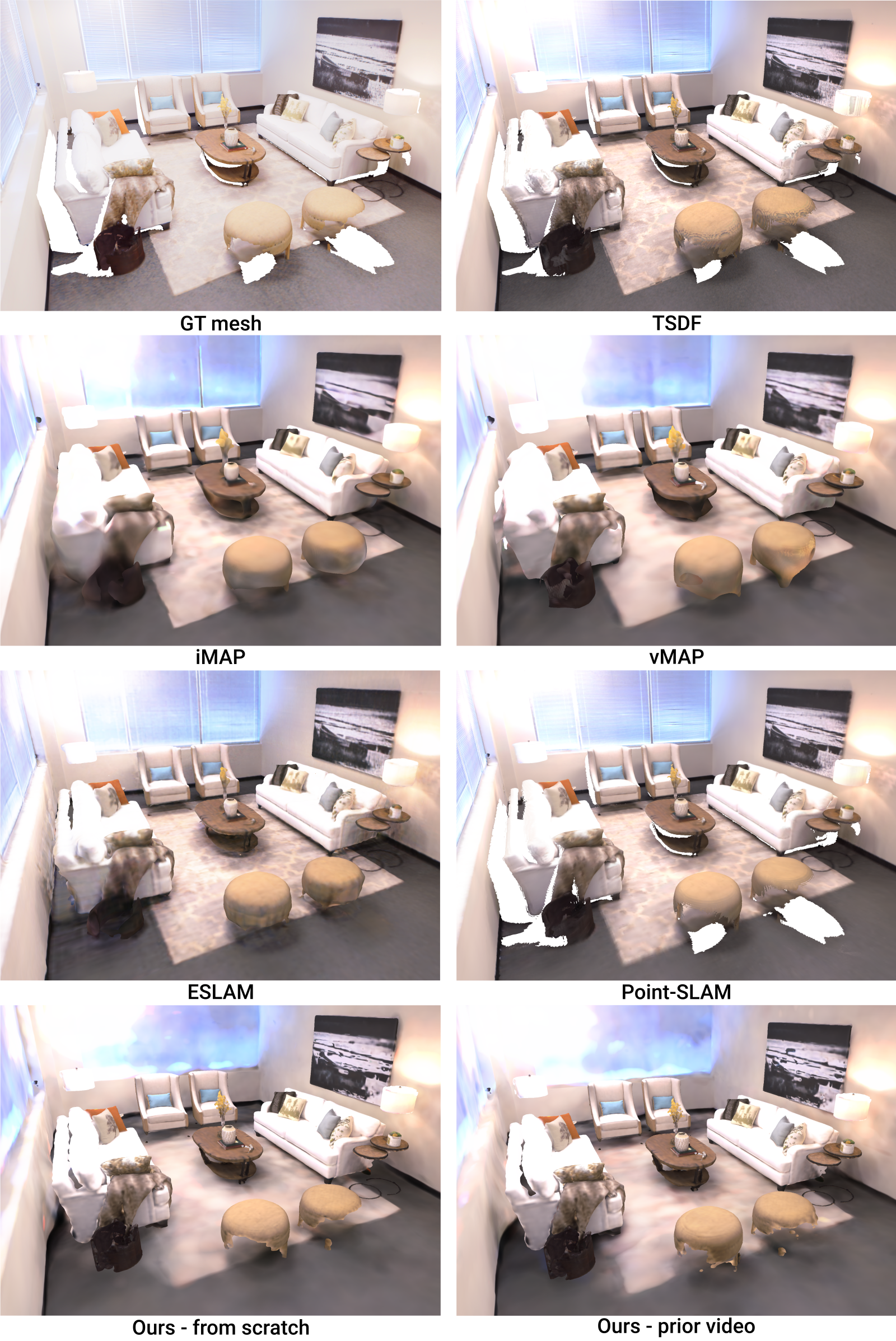}
    \vspace{-0.75em}
    \caption{\small{
    Visualization of reconstructed meshes at the scene level using TSDF-Fusion~\cite{curless1996tsdf}, iMAP~\cite{sucar2021imap} (through its reimplementation in~\cite{kong23vmap}), vMAP~\cite{kong23vmap}, ESLAM~\cite{johari2023eslam}, Point-SLAM~\cite{sandstrom2023pointslam} and our method, with or without object prior, on scene {\em room-0} of the Replica dataset~\cite{straub2019replica}.
    }}
    \label{fig:scene_level_room0}
\end{figure*}

We show meshes reconstructed by each of these methods on a Replica scene in~\Cref{fig:scene_level_room0}.
The ground truth mesh is displayed here only for the parts that are seen in the input sequence.
Methods relying on TSDF representations, \ie, TSDF~\cite{curless1996tsdf,zeng20173dmatch} and Point-SLAM for its mesh extraction~\cite{sandstrom2023pointslam}, are highly accurate for both objects and the background, though they do not reuse any prior information about the scene and are therefore unable to fill unseen parts.
iMAP~\cite{sucar2021imap} and vMAP~\cite{kong23vmap} extract oversmooth surfaces with some plausible completion for all objects and for the background, but they miss details of all reconstructed objects.
ESLAM, which relies on a single tri-plane representation for the whole scene, proposes some completion for unseen parts of objects and for the background, but it misses important details about thin structures like pouf feet or the basket on the ground (left of the image).
While our background model has limited ability to capture scene details, we obtain the highest level of details for all objects in the scene while being able to leverage prior knowledge to complete some parts, see for instance the back of poufs in the last image. 
We believe that further improvements in the background representation should make our method a strong competitor for online scene-level reconstructions.

\paragraph{\textbf{Additional visualizations on Replica.}}

\begin{figure*}[t]
    \centering
    \includegraphics[width=\linewidth]{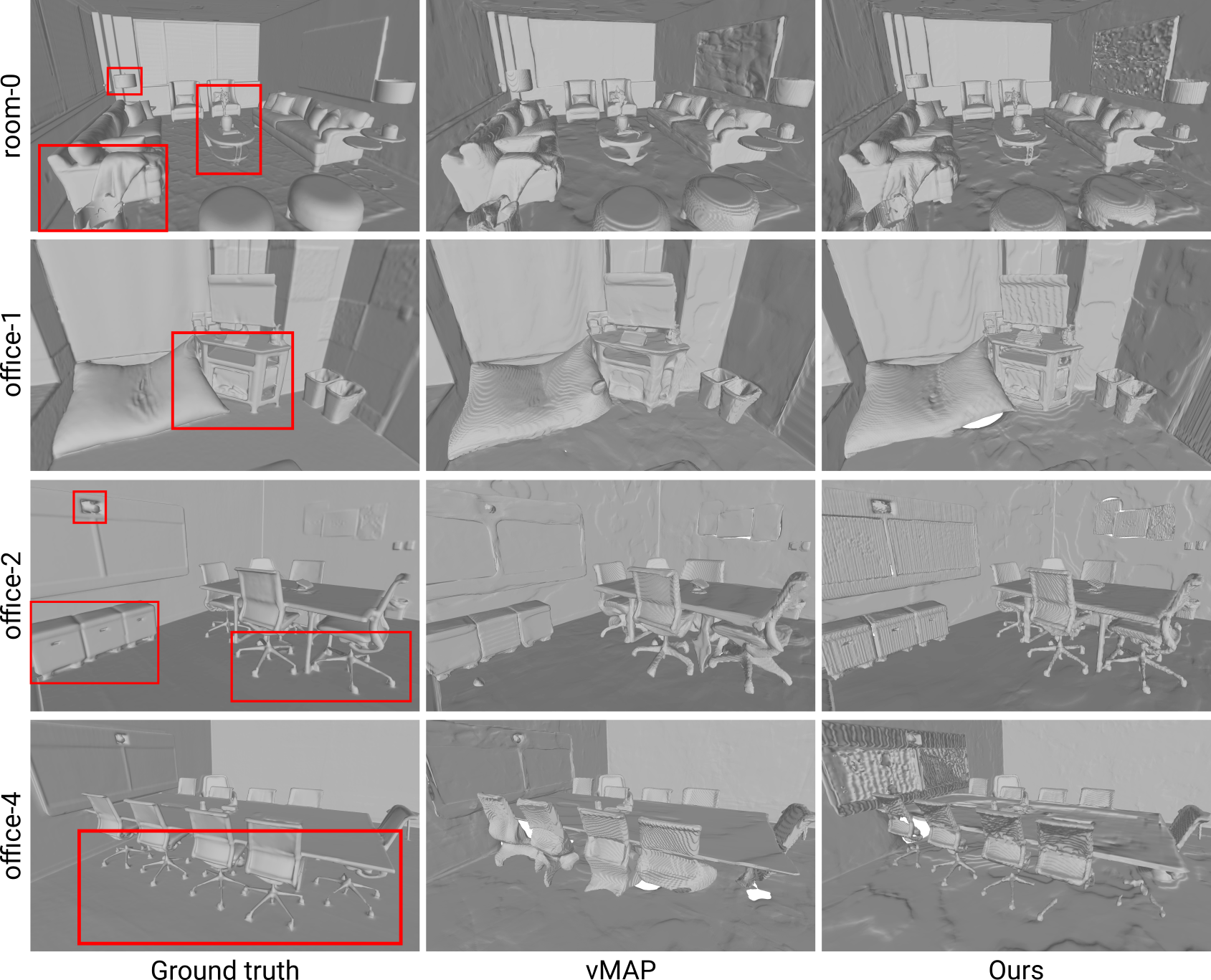}
    \caption{\small{
    Textureless meshes reconstructed with vMAP~\cite{kong23vmap} and our method on scenes of the Replica dataset~\cite{straub2019replica}.
    The textured version is presented in~\Cref{fig:replica-reconstructions} of the main paper.
    }}
    \label{fig:textureless_replica_meshes}
\end{figure*}

As textures may prevent the reader from observing the geometry details of a mesh, we provide a textureless version of~\Cref{fig:replica-reconstructions} from the main paper in~\cref{fig:textureless_replica_meshes}.
These textureless images emphasize on the higher level of details recovered by our method compared to vMAP~\cite{kong23vmap}.
In particular, thin structures, \eg, table and chair feet or handles, are more challenging for vMAP but are correcly recovered by our approach.

\paragraph{\textbf{Object meshes on ScanNet.}}
\begin{figure*}[t]
    \centering
    \includegraphics[width=\linewidth]{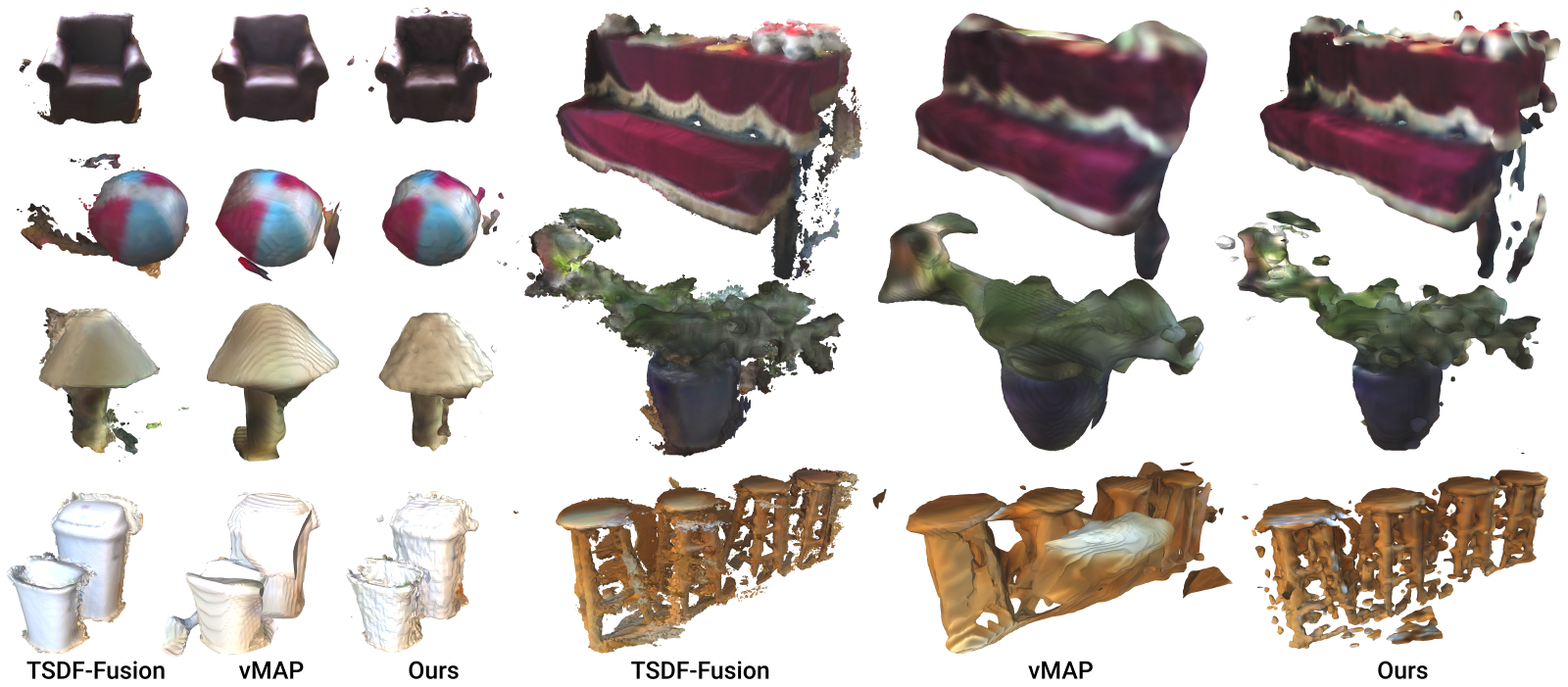}
    \caption{\small{
    Comparison of meshes between TSDF~\cite{curless1996tsdf,zeng20173dmatch}, vMAP~\cite{kong23vmap} and our method on objects from the ScanNet dataset~\cite{dai2017scannet}.
    Meshes are extracted with a resolution of 1cm.
    Our reconstructions are more accurate than vMAP while being more sensitive about depth and segmentation errors than TSDF.
    }}
    \label{fig:scannet_objects}
\end{figure*}

We show some reconstructed objects on the ScanNet dataset in \cref{fig:scannet_objects}, using our method as well as vMAP~\cite{kong23vmap} and a TSDF~\cite{curless1996tsdf, zeng20173dmatch} implemented with object grids of 5mm resolution.
All meshes for these visualizations are extracted at a 1cm resolution using marching cubes~\cite{lorensen1987marchingcubes}.
Unlike other methods, we provide TSDF with knowledge of each object extent before starting the reconstruction since it is a static representation.
Our method, run with no object prior, is able to reconstruct object geometries that are more accurate than vMAP~\cite{kong23vmap} on these real world sequences.
However, these sequences have very noisy depth and segmentation masks, resulting in artefacts in TSDF's reconstructions for ScanNet and slightly noisier ones for our method.
The update time for TSDF representations also grows significantly with the resolution of the grid and number of objects, making the access to finer reconstructions much more costly than coarse resolutions, unlike our representations which keep a constant computation speed for any object resolution.

\paragraph{\textbf{Additional results on sequences captured in our laboratory.}}

\begin{figure*}[t]
    \centering
    \includegraphics[width=\linewidth]{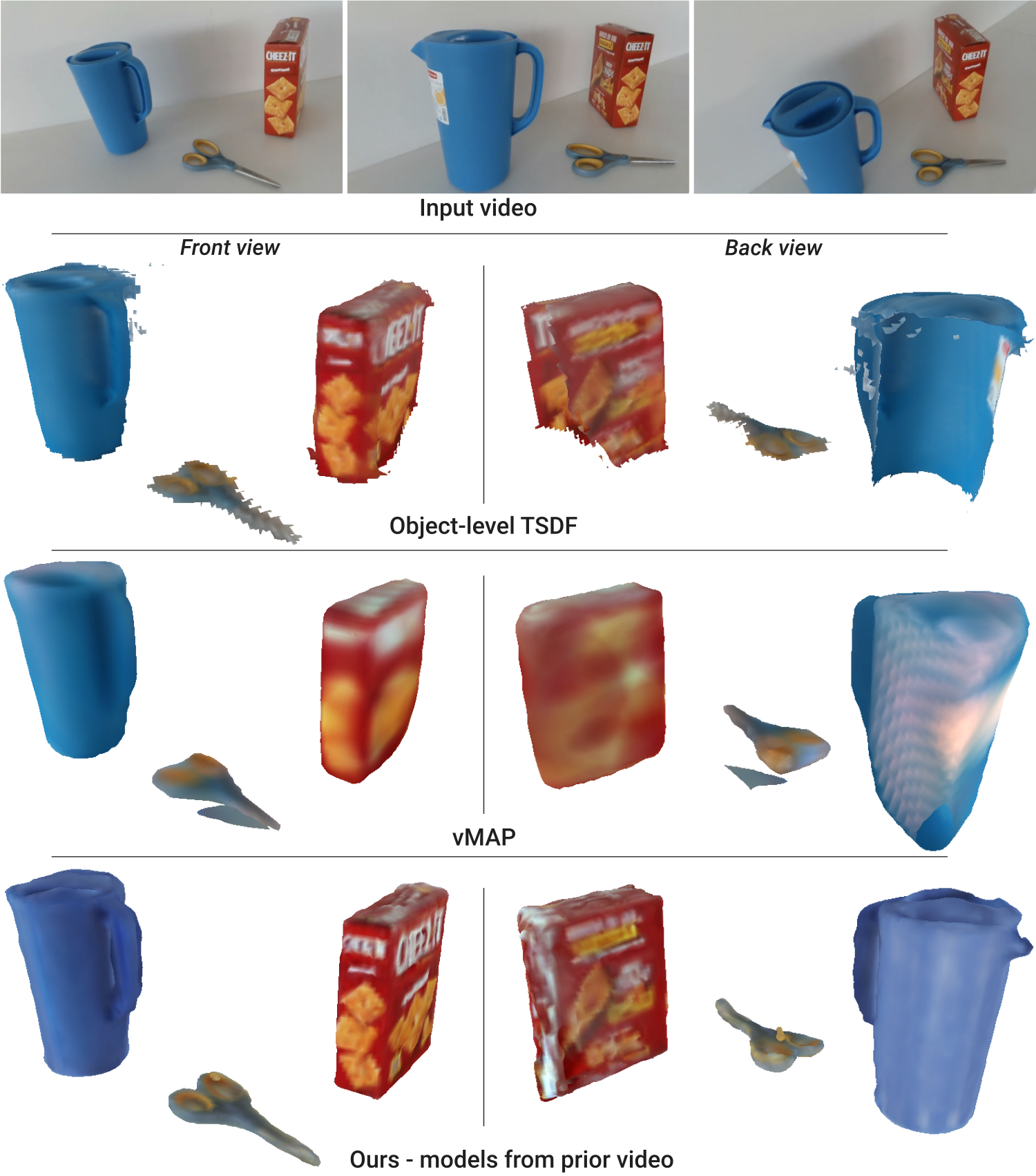}
    \vspace{-1.75em}
    \caption{\small{
    Additional reconstruction of a self-captured sequence with object-level TSDF, vMAP and our method using models from ground truth meshes, viewed from the front and the back.
    }}
    \label{fig:real_world_self_supp_mat_3objs}
    \vspace{-1.25em}
\end{figure*}

We show in~\Cref{fig:real_world_self_supp_mat_3objs} additional reconstructions on a scene with 3 objects, with views from the front and the back of the objects.
As in the main paper, TSDF reconstructs objects with some floating artefacts around their borders. This objects are incomplete for parts unseen in the current video.
vMAP outputs more complete but inaccurate object geometry with oversmoothed texture.
Conversely, our approach produces more faithful shapes and textures for both seen and unseen parts of objects, in particular thanks to object initializations from a prior video.

\paragraph{\textbf{Computation time.}}
On Replica’s room-0, our average times per frame are 740ms for objects reconstructed from scratch and 1.4s for models reused from the library.
On the same hardware, vMAP and iMAP take 420ms per frame, ESLAM takes 445ms, Point-SLAM needs 32s and the scene-level TSDF runs at 18ms per frame. 
Our implementation is however not yet parallelized, unlike vMAP, which would yield important time savings.
Important time savings can be obtained in several ways.
First, improving the implementation to parallelize the fitting of all object models instead of optimizing them sequentially should lead to significant speed gains.
Second, our retrieval and registration (R\&R) currently operate in the same thread as model optimization, which stops at each R\&R attempt, and require around 200ms per retrieval and registration attempt. 
Performing this stage in a separate thread should allow model optimization to run faster.
Third, as object models are fitted separately, their optimization can be paused after converging on the currently stored keyframes.
This is particularly useful for objects leaving the field of view for a long time, with no new stored keyframe.
In this case, fewer objects are optimized, which benefits both computation time and GPU memory.
Pausing the optimization is not feasible when considering the scene as a single entity.
Note that these times do not include segmentation and mask tracking times, which are assumed to be run separately.

\section{Implementation details}
\label{sec:implementation_details}

\subsection{Object-centric model and optimization}

\paragraph{\textbf{Sampling points.}}
At each frame and for each object, we perform 3 successive optimization steps, sampling 9600 rays among 6 keyframes for each step and 14 points per ray, 13 close to the surface and 1 closer to the camera.
As explained in Section 3.1 of the main paper, the surface sampling consists in drawing points close to the surface on rays $u$ following a normal distribution centered at the depth measurement $\mathcal{N}(D(u), \sigma)$. 
We take $3\sigma=5cm$ for Replica scenes and a larger $\sigma=10cm$ on ScanNet.
The latter was observed to give better reconstructions due to ScanNet's noisy depth measurements and outliers that grow excessively object bounding boxes, representing large empty spaces.

\paragraph{\textbf{Bounding boxes.}}
We compute our bounding boxes using axis-aligned boxes in the world frame defined for each scene. 
Such bounding boxes are more efficient to compute than randomly aligned ones, though they may encompass larger empty space.
When updating a bounding box, we add a 10\% margin to the box extent in order to avoid doing this update too often as parts of objects are discovered at each frame.
For each object, we only consider frames for which the object mask contains at least 100 pixels to avoid updating models based on too few observations.

\paragraph{\textbf{Keyframe criterion.}}
We reuse the same keyframe criterion as vMAP~\cite{kong23vmap}, which consists in considering every 25-th frame as a keyframe for objects and every 50-th for the background.
We store keyframes in a buffer of up to 20 keyframes for both objects and background.
Storing keyframes represents the largest memory usage of our approach, our object feature grids consisting in only around 65k parameters for shape and appearance respectively.

\paragraph{\textbf{Volume rendering details}}
We provide here more details about the rendering formulas and losses used for the online reconstruction.
For each ray $u$, we compute ray termination weights $w_{k,i}$ at each point as:
\begin{align}
    w_{k, i} &= \hat{o}_{k, i} \prod_{j=1}^{i-1} (1 - \hat{o}_{k, j}), 
    \label{eq:render-weight}
\end{align}
We then render the pixel color $\hat{C}_k$, depth $\hat{D}_k$, mask $\hat{M}_k$ and depth variance $\hat{V}_k$ of object $O_k$ as:
\begin{align}
    \hat{C}_k(u) = \sum_{i=1}^N w_{k,i} \hat{c}_{k, i},
    &\quad
    \hat{D}_k(u) = \sum_{i=1}^N w_{k,i} d_i,
    \label{eq:render-color-depth}
    \\
    \hat{M}_k(u) = \sum_{i=1}^N w_{k,i},
    &\quad
    \hat{V}_k(u) = \sum_{i=1}^N w_{k,i} (d_i - \hat{D}_k(u))^2. 
    \label{eq:render-mask-depth-uncertainty}
\end{align}
For each object $O_k$, the losses used during fitting penalize the difference between the inputs and the renderings:
\begin{align}
    \mathcal{L}_{col}(k, u) &= M_k(u) \|C(u) - \hat{C}_k(u)\|_1,
    \label{eq:loss_color}
    \\
    \mathcal{L}_{depth}(k, u) &= M_k(u) \frac{\|D(u) - \hat{D}_k(u)\|_1}{\sqrt{\hat{V}_k(u)}},
    \label{eq:loss_depth}
    \\
    \mathcal{L}_{mask}(k, u) &= \|M_k(u) - \hat{M}_k(u)\|_1,
    \label{eq:loss_mask}
\end{align}
where $M_k$ is the binary mask of $O_k$.

\paragraph{\textbf{Optimization details.}}
We implement our feature grids and MLP using the tiny-cuda-nn library~\cite{tiny-cuda-nn}. 
Our object models are optimized with AdamW~\cite{loschilov2017adamw} with learning rates $5 \times 10^{-3}$ and $3.5 \times 10^{-4}$ respectively for the feature grids and MLPs, and weight decay $0.1$ for both.
For the background model, we reuse the same parameters as the vMAP paper~\cite{kong23vmap}.

\subsection{Integrating object shape priors}

\paragraph{\textbf{Constructing the object library.}}
\label{par:object_library}
\begin{figure*}[t]
    \centering
    \includegraphics[width=\linewidth]{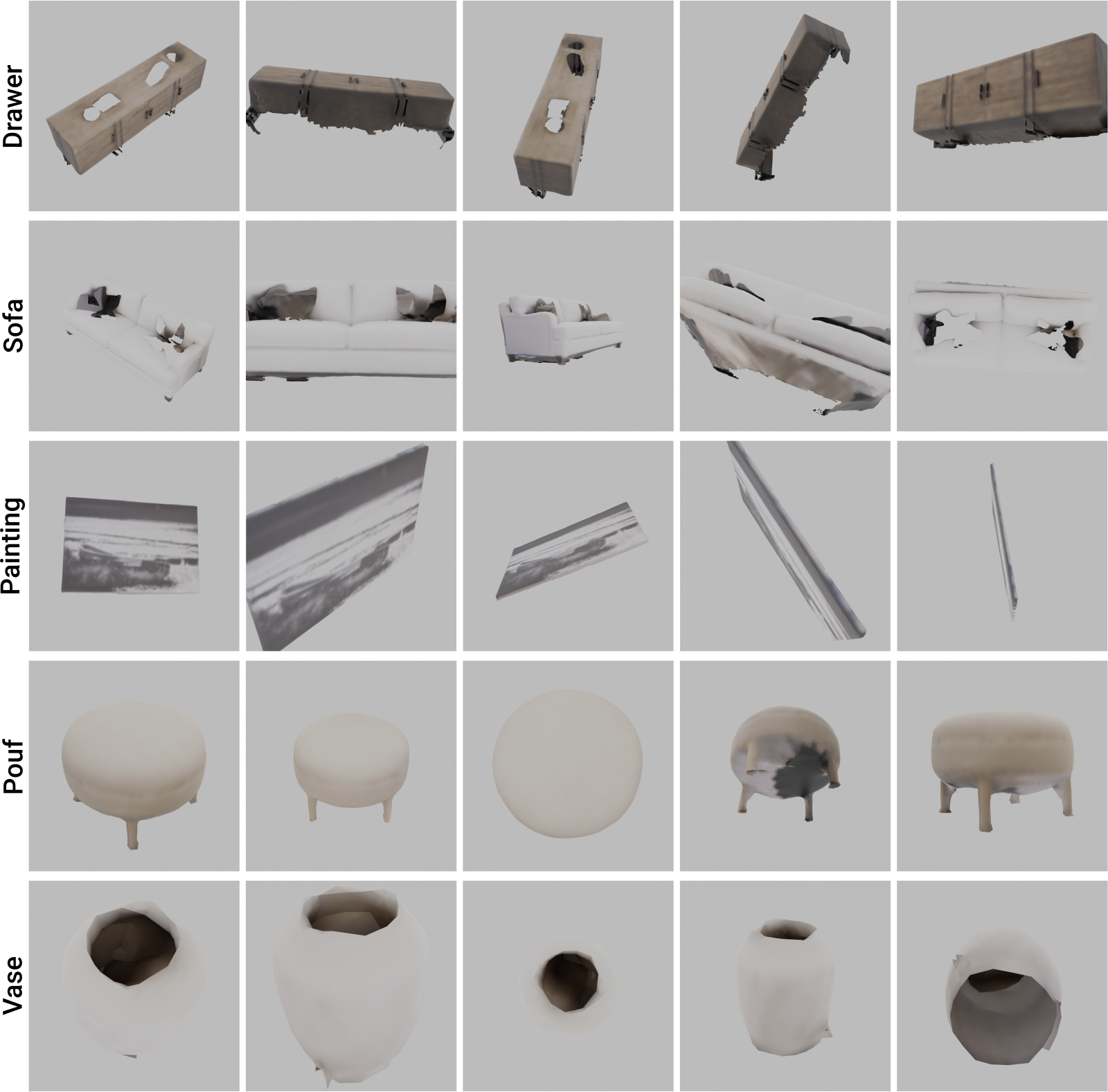}
    \caption{\small{
    Examples of rendered images of ground truth meshes using the Blenderproc~\cite{Denninger2023blenderproc} renderer.
    Our object models that leverage prior knowledge of GT meshes are fitted on these images.
    }}
    \label{fig:renders}
\end{figure*}

As explained in~\Cref{subsec:init-database} of the main paper, we build object models offline from either full 3D meshes or video sequences.
We detail here the first case.
For each object 3D mesh, we render 40 images from random viewpoints around the object using the BlenderProc~\cite{Denninger2023blenderproc} renderer, each image being of size $1024\times1024$.
We show examples of these renders in~\cref{fig:renders}.
3D meshes for the Replica dataset are ground-truth object meshes and have been obtained by extracting a closed surface from a single volumetric representation.
The latter extraction of object meshes performed by vMAP~\cite{kong23vmap} consists in splitting this closed surface in objects according to vertex instance Ids, resulting in all objects being open surfaces.
Thus, if camera poses are randomly sampled all around an object for our renders, the same part of a surface may be observed from two opposite viewpoints, which in turn causes problems during reconstruction.
To avoid that issue, we use normal information to only retain one side of the surface.
From these rendered images, we fit an object model with the method explained in~\Cref{subsec:object-centric} of the main paper, with the only difference that all inputs are known at the beginning of the fitting. 
Hence, we do not need to perform the online optimization but instead sample all frames at each optimization step.
We perform 500 optimization steps per object and store the model at the last step for our database.

\paragraph{\textbf{Retrieval and registration.}}
For retrieval, we use the CLIP~\cite{radford2021clip} version {\em ViT-bigG-14} from OpenClip~\cite{ilharco2021openclip,cherti2022reproducibleclip} and filter out retrieved objects for which the cosine similarity score is larger than $0.7$.
For the registration part, the FPFH~\cite{rusu2009fpfh} features, Ransac~\cite{fischler1981ransac} and point-to-plane ICP~\cite{chen1992icppoint2plane} algorithms are reused from Open3D's implementation~\cite{zhou2013open3d}.
We filter the fitness with a threshold of $0.8$ and keep registered poses for which at least $90\%$ of reprojected points in the camera frame belong to the input object mask and have a depth larger than the depth measurement, with a tolerance of $2cm$.

\paragraph{\textbf{Synthesizing keyframes on the fly.}}
Once an object model has been initialized from the object library, we use the retrieved model to render additional views to fit the current object model.
In this way, we sample half of the camera poses at each optimization step among the poses stored in the object library.
We then render color, depth and mask using 24 points per ray, which we sample uniformly in the whole bounding box. 
This sampling contains more points than the one from current views (\ie, 14 points) to cope with the absence of depth information that would otherwise guide the sampling around the actual object surface.

\subsection{Evaluation datasets}
\begin{figure*}[t]
    \centering
    \includegraphics[width=\linewidth]{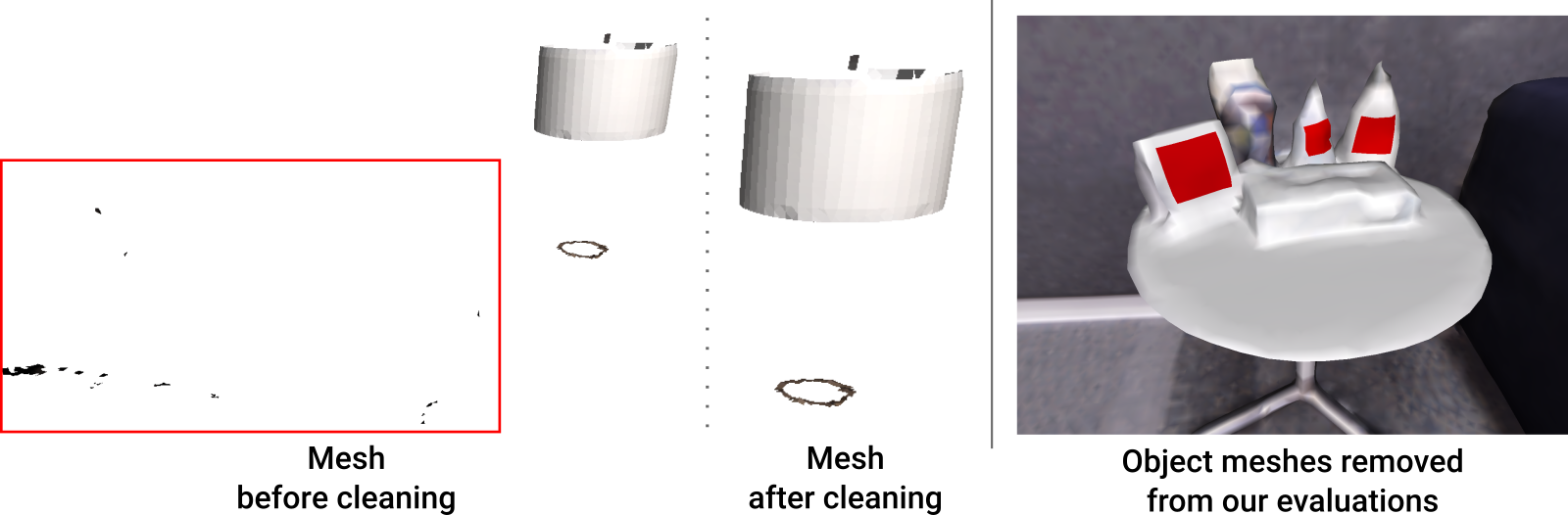}
    \caption{
    \small{
    Examples of meshes cleaned before running our evaluations in the main paper.
    {\em (Left)}: few objects are cleaned to remove outliers, highlighted in red here.
    {\em (Right)}: other small and flat objects like product tags, colored in red, are also discarded.
    }
    }
    \label{fig:cleaned_meshes}
\end{figure*}

\paragraph{\textbf{Replica.}}
Objects considered for the evaluations of the main paper slightly differ from those used in vMAP~\cite{kong23vmap}.
We clean few noisy meshes to remove vertices that are outliers and discard a few other tiny objects, \ie, with fewer than 50 vertices.
Examples of such objects are shown on \Cref{fig:cleaned_meshes}.
Following this cleaning, Replica scenes contain on average 50 objects each.

\paragraph{\textbf{ScanNet.}}
For the ScanNet dataset, we additionally pre-process each depth image to remove outliers.
More specifically, at each frame and for each object $O_k$, we compute the mean $m_k$ and standard deviation $s_k$ of depth values falling in the object mask and discard points whose depth is outside the range $[m_k-\alpha s_k, m_k+\alpha s_k]$, where we choose $\alpha=1.5$, making this interval close to the 90\% confidence interval of normal distributions.
We also compute a histogram of depth points belonging to mask $k$ with 15 values in the camera depth range ($[0m, 6m]$) and keep only bins that contain at least $5\%$ of points.
This removes a large number of outliers, though some remain that may have a strong impact on our object bounding boxes.
Further joint pre-processing of depth maps and segmentation masks would benefit our reconstructions on real world images.
For fair comparison in the real world reconstructions, we also apply this pre-processing to TSDF~\cite{curless1996tsdf,zeng20173dmatch} and vMAP~\cite{kong23vmap}.

\paragraph{\textbf{Real-world sequences.}}
For our videos, we apply the same depth image processing as for ScanNet sequences and additionally erode object masks by few pixels to remove outliers and obtain better geometry.

\subsection{Evaluation metrics}
For evaluation on seen parts of meshes, we first cull vertices that are not seen in any input frame.
We reuse ESLAM's culling script~\cite{johari2023eslam} and remove points at a depth $D_{rec} > D_{input} + \tau$, where $D_{rec}$ is the depth of reprojected mesh points in camera frames, $D_{input}$ is the measured depth for that frame and $\tau$ is a tolerance.
We choose $\tau=3cm$ for the scene meshes and $2cm$ for object meshes which we found to be a good trade-off between keeping all seen reconstructed vertices that may be inaccurately positioned and discarding all unseen ones.
When evaluating reconstructions on the whole objects, including parts that are not seen in the input video sequence, we consider the full ground truth meshes, without applying this culling operation.

\clearpage

\end{document}